\documentclass[review,onefignum,onetabnum]{siamart171218}

\usepackage{lipsum}
\usepackage{amsfonts}
\usepackage{graphicx}
\usepackage{epstopdf}
\usepackage{algorithmic}

\ifpdf
  \DeclareGraphicsExtensions{.eps,.pdf,.png,.jpg}
\else
  \DeclareGraphicsExtensions{.eps}
\fi

\newsiamremark{remark}{Remark}
\newsiamremark{hypothesis}{Hypothesis}
\crefname{hypothesis}{Hypothesis}{Hypotheses}
\newsiamthm{claim}{Claim}

\headers{Initial condensation for three-layer NN}{Zheng-an chen and tao luo}

\title{On the dynamics of three-layer neural networks: initial condensation\thanks{Submitted to the editors DATE.
}}

\author{Zheng-an Chen \thanks{School of Mathematical Sciences, and CMA-Shanghai, Shanghai Jiao Tong University (\email{zhengan$\_$chen@sjtu.edu.cn}).}
\and Tao Luo \thanks{Corresponding author. School of Mathematical Sciences, CMA-shanghai, Institute of Natural Sciences, MOE-LSC, Shanghai Jiao Tong University, and Shanghai Artificial Intelligence Laboratory(\email{luotao41@sjtu.edu.cn}}).
}

\usepackage{amsopn}

\makeatletter
\newcommand*{\addFileDependency}[1]{
  \typeout{(#1)}
  \@addtofilelist{#1}
  \IfFileExists{#1}{}{\typeout{No file #1.}}
}
\makeatother

\newcommand*{\myexternaldocument}[1]{%
    \externaldocument{#1}%
    \addFileDependency{#1.tex}%
    \addFileDependency{#1.aux}%
}

\usepackage{mlmath}
\usepackage{graphicx}
\usepackage{subcaption}
\usepackage{enumitem}

\ifpdf
\hypersetup{
  pdftitle={Initial condensation for three-layer NN},
  pdfauthor={TBD }
}
\fi

\myexternaldocument{ex_supplement}

\begin{document}

\maketitle

\begin{abstract}
Empirical and theoretical works show that the input weights of two-layer neural networks, when initialized with small values, converge towards isolated orientations. This phenomenon, referred to as condensation, indicates that the gradient descent methods tend to spontaneously reduce the complexity of neural networks during the training process. In this work, we elucidate the mechanisms behind the condensation phenomena occurring in the training of three-layer neural networks and distinguish it from the training of two-layer neural networks. Through rigorous theoretical analysis, we establish the blow-up property of effective dynamics and present a sufficient condition for the occurrence of condensation, findings that are substantiated by experimental results. Additionally, we explore the association between condensation and the low-rank bias observed in deep matrix factorization.
\end{abstract}



\section{Introduction}\label{sec::Intro}
In deep learning, different parameter initialization methods cause distinct training dynamics and generalization behavior, which is  regarded as an important part of implicit regularization of neural networks~(NNs). Specifically, training dynamics of NNs with large initialization can be approximated by a kernel regression known as neural tangent kernel (NTK)\cite{jacot2018neural,huang2020dynamics,arora2019exact} and the lazy training has been characterized in \cite{chizat2019lazy} by which NN parameters stay close to
initialization during the training. Oppositely, small initialization often lead to parameter condensation~\cite{luo2021phase,xu2021towards,chen2023phase}, where NNs tend to represent target function with fewer effective neurons. Fall in the between, gradient flow (GF) of NNs can be captured by a nonlinear partial differential equation through a mean field viewpoint\cite{mei2018mean,sirignano2020mean,rotskoff2018parameters,chizat2018global}.

In category of two layer NNs, \cite{luo2021phase} establish a phase diagram by finding appropriate to bridging dynamics induced by different initialization. Because of highly nonlinear structure, the study of multi-layer NNs is more challenging. Three-layer NN is a good starting point as interaction between parameters is relatively simple.  An empirical phase diagram established in \cite{zhou2022empirical}  shows complicated dynamical regimes exhibited by three-layer NNs, which reveals significant difference with two-layer NNs.  

In this work, we focus on condensed regime associated with small initialization in three-layer NNs. The significance of this work lies in two aspects. On one hand, condensed regime reveals the possibility that narrow NNs can be embedded into wider NNs which revealed by embedding principle \cite{zhang2021embedding,zhang2021embeddingnips}, which means complexity of NNs as output function is much fewer than the dimension of parameter space. It is believed that such NNs have good generalization capabilities due to the well known bound of generalization error by complexity \cite{bartlett2002rademacher}.  On the other hand, this work will provide valuable insights to differentiate the optimization dynamics between three-layer and two-layer networks. Our contribution can be summarized as follows:
\begin{enumerate}
    \item We model the asymptotic dynamics of gradient descent in three-layer NNs and prove blow-up property of the effective model (Theorem \ref{thm::BlowUp}), which is significantly different from the situation in two-layer networks.
    \item We propose Assumption \ref{assump::FinalStageCond} and reveal its link to condensation and effectiveness by experimental  approaches (Figure \ref{fig::ExperiCondensation} and \ref{fig::ExperimentFSC}).
    \item We prove effective dynamics has condensed solution under Assumption \ref{assump::FinalStageCond} (Theorems \ref{thm::UpperBound} and \ref{thm::Condense}).
\end{enumerate}

The organization of the paper is listed as follows. In Section 2, we discuss some
related works. In Section 3, we give some preliminary introduction and basic properties that are different from the dynamics of two-layer networks. In Section 4, we state our main results through experimental and theoretical approaches. Some discussions of the connection between initial condensation and matrix factorization are deferred to the Appendix.

\section{Related Works}\label{sec::RelaWork}
The investigation into the dynamics of parameter condensation was initiated by Luo et al. \cite{luo2021phase}. In their seminal work, the team utilized conservation laws and proof by contradiction  to quantify the attributes linked to the condensation phenomenon. At that stage, insights into the parameter directionality were non-existent. Zhou et al. \cite{zhou2022towards} subsequently made strides in this area, computing the target direction through a fixed-point algorithm. However, their analysis remained largely empirical, constrained by the absence of a rigorous error estimation that would reconcile the linearized and the genuine dynamic scenarios. This gap in the literature was addressed by Chen et al. \cite{chen2023phase}, who refined the understanding of parameter condensation by providing a detailed analysis within the framework of two-layer NNs.

Maennel et al. \cite{maennel2018gradient} delve into the phenomena characteristic of the condensed regime in two-layer ReLU NNs. Their investigation reveals that as the parameter initialization nears the zero vector, a distinctive quantization effect emerges. This effect denotes a tendency for the weight vectors to align with a finite set of orientations. Intriguingly, these orientations are predominantly influenced by the input data during the initial phase of the training process.


The methodology primarily involves examining the dominant dynamics, which closely aligns the model with the behavior of linear networks. Linear networks are a particular subset of fully connected NNs, with matrix factorization being their most thoroughly investigated application. In this context, linear networks are typically trained to approximate a low-rank matrix. Within this body of research, a pivotal scaling has been discerned that delineates different operational regimes in matrix factorization, paralleling those found in supervised learning \cite{woodworth2020kernel}.
Studies by \cite{arora2019implicit,gunasekar2017implicit} delve into the implicit regularization effects inherent in gradient descent when initiating training with a 'balanced initialization'. Employing such an initialization, we are able to derive some of our findings through similar reasoning. Consequently, our research serves to broaden the scope of these preceding analyses.

\section{Preliminaries}\label{sec::Prel}
\subsection{Notations}\label{sec::Nota}
First, we introduce some notations that will be used in the rest of this paper. Let $n$ and $m$ be  the number of samples and  the width of hidden layers, respectively. Let $[n] $ denote the set of integers from $1$ to $n$. Denote vector $L^2$ norm as $\norm{\cdot}_2$ and matrix Frobenius norm as $\norm{\cdot}_{\mathrm{F}}$. Let $\langle \cdot, \cdot \rangle$ represent standard inner product between two vectors. For a vector $\bm{v}$, denote its $k$-th entry as $\bm{v}_k$. For a matrix $\vA$,  denote the element in the $k$-th row and $k'$-th column as $\vA_{k k'}$. And denote $k$-th row as $\vA_k$ and $k'$-th column as $\vA^{k'}$. Unless otherwise specified, summation `$\sum$' is performed over the network width.
\subsection{Problem Formulation}\label{sec::ProbForm}

We initiate the modeling of gradient descent dynamics for three-layer NNs through  asymptotic analysis method. Under very mild assumptions, we demonstrate that this effective dynamics of three-layer NNs undergoes finite-time blow-up, which is essentially different from those of two-layer NNs.

Considering a three-layer neural network
\begin{equation}
    f_{\vtheta}(\vx) = {\va}^{\T} \sigma \left( {\mW}^{[2]} \sigma \left( {\mW}^{[1]} \vx \right)  \right),
\end{equation}
where $\va \in  {\sR}^m $, ${\mW}^{[2]} \in {\sR}^{m \times m}$ , ${\mW}^{[1]} \in {\sR}^{m \times d}$ and $\sigma$ is any smooth activation function with $\sigma(0)=0$. Its parameters are denoted  as  $\vtheta= \text{vec} \{ \va , \mW^{[2]} ,\mW^{[1]} \}$ by flattening $\va$, $\vW^{[1]}$ and $\vW^{[2]}$ and concatenating them together.  The empirical risk  for sample set $\vS =\{ (\vx_i,y_i)\}_{i=1}^n$ to be minimized is given by 
\begin{equation}
    {\mR}_{\mS}(\vtheta)=\frac{1}{2n} \sum_{i=1}^n (f_{\vtheta}({\vx}_i)-y_i)^2 .
\end{equation}

We use gradient descent method (GD) to optimize the empirical risk. The parameters are initialized as follows:
\begin{equation}
    a_k \sim \fN (0,\varepsilon^2) ,  {\mW}_{kk'}^{[2]} \sim \fN (0,\varepsilon^2), {\mW}_{kk'}^{[1]} \sim \fN (0,\varepsilon^2), 
\end{equation}
where $\varepsilon$ is a small parameter. Usually, people use gradient flow (GF) which is continuous limit  of gradient descent method to analyze the training dynamics.

Normalizing the parameters by 
\begin{equation*}
    \bar{\va}=\varepsilon^{-1} \va , \bar{\mW}^{[2]}= \varepsilon^{-1} \mW^{[2]}, \bar{\mW}^{[1]}= \varepsilon^{-1} \mW^{[1]}, \bar{\vtheta}= \varepsilon^{-1} \vtheta,
\end{equation*}
we have power series expansion of empirical risk in terms of  $\varepsilon$ as follows:
\begin{equation}
    \mR_{\mS} (\vtheta)= \frac{1}{2n} \sum_{i=1}^n y_i^2- \varepsilon^3 \bar{\va}^{\T} \bar{\mW}^{[2]} \bar{\mW}^{[1]} \frac{(\sigma'(0))^2}{n} \sum_{i=1}^n y_i {\vx}_i + o(\varepsilon^3).
\end{equation}
Then the leading order of GF obeys the following dynamics
\begin{equation}
    \varepsilon \frac{\D \bar{\vtheta}}{ \D t} =\varepsilon^2 \nabla_{\bar{\vtheta}} \bar{\va}^{\T} \bar{\mW}^{[2]} \bar{\mW}^{[1]} \frac{(\sigma'(0))^2}{n} \sum_{i=1}^n y_i {\vx}_i.
\end{equation}
Rescaling the time by setting
\begin{equation*}
    \bar{t}= \varepsilon \frac{(\sigma'(0))^2}{n} \Norm{ \sum_{i=1}^n y_i \vx_i }_2  t
\end{equation*}
and dropping the bar in the above dynamics for simplicity, we get the effective model that will be mainly discussed through this paper 
\begin{equation}
    \frac{\D \vtheta}{\D t}= \nabla_{\vtheta} \va^{\T} \mW^{[2]} \mW^{[1]} \bm{v},
\end{equation}
where 
\begin{equation}
    \bm{v}=\frac{\sum_{i=1}^n y_i {\vx}_i}{\| \sum_{i=1}^n y_i \vx_i \|}_2
\end{equation}
is called the target direction. Obviously, $\bm{v}$ only depends on samples.

We further introduce energy $E$ as 
\begin{equation}
    E=\va^{\T} \mW^{[2]} \mW^{[1]} \bm{v},
\end{equation}
based on which the effective model can be regarded as the gradient ascent dynamics of the energy.
More precisely, that is 
\begin{equation}
    \left\{
    \begin{aligned}
    &\frac{\D \va}{\D t}=\mW^{[2]} \mW^{[1]} \bm{v},\\
    &\frac{\D \mW^{[2]}}{\D t}=\va(\mW^{[1]} \bm{v})^{\T},\\
    &\frac{\D \mW^{[1]}}{\D t}={\mW^{[2]}}^{\T} \va \bm{v}^{\T}.
    \end{aligned}
   \right.
\end{equation}

To simplify the notation, we set $b_{kk'}=\mW^{[2]}_{kk'}$ and $c_{k'}=(\mW^{[1]} \bm{v})_k$. Thus, the dynamics simply reads as
\begin{equation}
\label{eq::SimpEffectiveModel}
     \left\{
    \begin{aligned}
    &\frac{\D \va}{\D t}=\vb \vc,\\
    &\frac{\D \vb}{\D t}=\va \vc^{\T},\\
    &\frac{\D \vc}{\D t}=\vb^{\T} \va.
    \end{aligned}
   \right.
\end{equation}
In the rest of the paper, we will first investigate the general properties of Equation \eqref{eq::SimpEffectiveModel} and then characterize condensation based on such properties. See the details in Section \ref{sec::Intro}.


\subsection{Blow Up}\label{sec::BlowUp}

First, we have the following proposition by direct computation. The proposition illustrates the symmetry of the solution and elucidates the interplay between $\va$,$\vc$, and $\vb$.

\begin{prop}[conservation laws]\label{prop::ConservationLaw}
Dynamical system \eqref{eq::SimpEffectiveModel} satisfies following equalities.
\begin{equation}
    \left\{
    \begin{aligned}
    &\frac{\D}{\D t}  a_k^2=\frac{\D}{\D t}  \sum_{k'} b_{kk'}^2,\\
    &\frac{\D}{\D t}  c_k^2=\frac{\D}{\D t}  \sum_{k}  b_{kk'}^2,\\
    &\frac{\D}{\D t} \| \va \|_2^2=\frac{\D}{\D t} \|\vb\|_{\mathrm{F}}^2=\frac{\D}{\D t} \| \vc \|_2^2=2E.
    \end{aligned}
   \right.
\end{equation}
\end{prop}
\begin{proof}
    We just prove for $\va$ and $\vb$ since the system is symmetric. We have 
    \begin{equation*}
        \frac{\D}{\D t} a_k^2  = 2 \sum_{k'} a_k b_{kk'}c_{k'} 
                               = \frac{\D}{\D t} \sum_{k'} b_{kk'}^2.
    \end{equation*}
    This finished the proof of  first two equation. Then we take sum for $k$, we have 
    \begin{equation*}
            \frac{\D}{\D t} \norm{\va}_2^2  = \frac{\D}{\D t} \sum_{k} a_k^2 
                                            = \sum_{k,k'}a_k b_{kk'}c_{k'} 
                                            = 2E.
    \end{equation*}
    This is just the third equation.
\end{proof}

Next we introduce the following assumption.
\begin{assump}[blow up initial data]
\label{assump::BlowUpIniData}
The initial norm of $\va$ and $\vc$ is not equal. That is 
\begin{equation}
    \norm{\va(0)}_2 \neq \norm{\vc(0)}_2.
\end{equation}
We also assume that 
\begin{equation}
\left\{
    \begin{aligned}
        \| \dot{\va}(0) \|_2^2 - \| \dot{\vc}(0) \|_2^2 -\|\vc(0)\|_2^2 ( \|\vc(0) \|_2^2 -\| \va(0) \|_2^2) \neq 0 &,   \|\va(0)\|_2 > \|\vc(0)\|_2, \\
        \| \dot{\va}(0) \|_2^2 - \| \dot{\vc}(0) \|_2^2 -\|\va(0)\|_2^2 ( \|\vc(0) \|_2^2 -\| \va (0) \|_2^2) \neq 0 &, \|\va(0)\|_2 < \|\vc(0)\|_2. \\
    \end{aligned}
\right.
\end{equation}    
\end{assump}
\begin{rmk}
Assumption \ref{assump::BlowUpIniData} serves as a sufficient condition for the blow-up property. Despite the peculiar form of the inequality constraint, this event always occurs with probability $1$. Thus, the assumption is not restrictive.
\end{rmk}
For completeness, we provide a particular initial value problem to illustrate that the above assumption is also necessary.
\begin{exam}
Consider dynamical system \eqref{eq::SimpEffectiveModel} with initial condition:
$$
\va(0) = (1,0)^{\T} , 
\vb(0) = \begin{pmatrix}
1 & 0 \\
0 & 0 \\
\end{pmatrix},
\vc(0) = (-1,0)^{\T}.
$$
On one hand,  $\norm{\va(0)}_2 = \norm{\vc(0)}_2=1$, and thus this example does not satisfy Assumption \ref{assump::BlowUpIniData}.
On the other hand, this example satisfies balanced initialization. That is 
\begin{equation*}
    \va(0) \va^{\T}(0) = \vb(0) \vb(0)^{\tT} = \vc(0)\vc(0)^{\T}.
\end{equation*}
Thus energy $E$ satisfies following equation based on Theorem 1 in \cite{arora2018optimization}:
\begin{equation*}
    \dot{E} = 3 E^{\frac{4}{3}}
\end{equation*}
This implies that energy $E$ converges to 0.  
Therefore, without Assumption \ref{assump::BlowUpIniData}, dynamical system may not blow up. 
\end{exam}

Based on Assumption \ref{assump::BlowUpIniData}, we have the following theorem which describes the property of finite-time blow-up.
\begin{thm}[blow up]
\label{thm::BlowUp}
Suppose that Assumption \ref{assump::BlowUpIniData} holds. Consider dynamical system \eqref{eq::SimpEffectiveModel}. There exists a finite time $T^*>0$ such that $\lim_{t \rightarrow T^*} E(t)=+ \infty$.
\end{thm}

\begin{proof}
By local Lipshcitz condition on the right hand side of dynamical system \eqref{eq::SimpEffectiveModel}, it  has a solution for $t \in (0,T^*)$ where $T^*$ is maximum existence time of solution and can be  infinity. Taking derivative of $E$, we have 
\begin{equation}
\label{eq::DeriE}
    \dot{E}=\frac{\D}{\D t}(\va^\T \vb \vc ) 
    =\| \dot{\va} \|_2^2 +\| \dot{\vc} \|_2^2 +\| \va \|_2^2 \|\vc \|_2^2 .
\end{equation}
The inequality of arithmetic and geometric means leads to  
\begin{equation*}
\label{eq::IneqDeriE}
    \begin{aligned}
       \dot{E} & \ge 3 (\| \dot{\va} \|_2^2 \| \dot{\vc} \|_2^2  \| \va \|_2^2 \|\vc \|_2^2)^{\frac{1}{3}} \\
    & \ge [(\dot{\va}^{\T} \va)^2 (\dot{\vc}^{\T} \vc)^2]^ {\frac{1}{3} } \\
    &= 3E^{\frac{4}{3}}.
    \end{aligned}
\end{equation*}
It implies that energy $E$ increase monotonically. If  $E(0)>0$, 
\begin{equation*}
    \frac{\D}{\D  t} E^{-\frac{1}{3}} \le -1.
\end{equation*}
Integrating both sides of the inequality, we have a lower bound of energy $E$
\begin{equation}
\label{eq::LowBd}
    E(t) \ge  \frac{1}{(E(0)^{-\frac{1}{3}} -t)^3}.
\end{equation}
Thus, in the case where $E(0)>0$,we have $T^* \le E(0)^{-\frac{1}{3}}$.

For the case $E(0) \le 0$, we consider $-E(t)$ and get 
\begin{equation}
\label{eq::SimpleUpperBound}
    E(t) \ge - \frac{1}{(t + (-E(0))^{-\frac{1}{3}})^3}.
\end{equation}

We claim that there exists some time $t_0 >0$, such that $E(t_0)>0$. We prove this claim by the method contradiction.

Suppose that $E(t) \le 0$ for all $0<t<T^*$. Recall that $\dot{E}(t) \ge 0$ for all $0<t<T^*$. The boundedness of $\va$, $\vb$, and $\vc$, as well as the monotonicity of energy $E$,  implies that $E(T^*)=\lim_{t \rightarrow T^*}  E(t)$ exists and $-\infty < E(T^*) \le 0$.  We discuss in different situations.

(\romannumeral 1) The case of  $T^*< + \infty $.  Solutions can be extended to a time larger than $T^*$ since  $\norm{\va}_2, \norm{\vb}_{\text{F}} , \norm{\vc}_2$ are bounded due to the conservation law. This contradicts with the definition of $T^*$. 

(\romannumeral 2) The case of  $T^* = + \infty$ and $E(T^*)<0$. That is $\lim_{t \rightarrow +\infty} E (t) <0$. However, this contradicts with Equation \eqref{eq::SimpleUpperBound}. 


(\romannumeral 3)  The case of  $T^* = + \infty$ and $E(T^*)=0$. That is $\lim_{t \rightarrow +\infty} E (t) =0$. We prove this case in three steps.

\textbf{Step 1}: We show $\lim_{t \rightarrow \infty} \| \va(t)\|_2^2 \| \vc(t)\|_2^2 =0$. Since $E(t) \le 0$ for all $t$, $\|\va\|_2^2$, $ \|\vb\|_{\mathrm{F}}^2 $, $\| \vc \|_2^2 $ decreases monotonically. However, they are bounded below by $0$. Hence, they have limits and are always bounded in this case. Also, we note that $\dot{E} \ge 0$.  If $\liminf_{t \rightarrow \infty} \dot{E}(t) >0$, it contradicts with the fact that $\lim_{t \rightarrow \infty} E(t)= 0$. This implies 
\begin{equation*}
    \lim_{t \rightarrow \infty} \| \va(t)\|_2^2 \| \vc(t)\|_2^2 =0.
\end{equation*}
Thus we have either $\lim_{t \rightarrow \infty} \|\va(t)\|_2^2 =0 $ or $\lim_{t \rightarrow \infty} \|\vc(t)\|_2^2 =0$. 
 We can also easily get  $\lim_{t \rightarrow \infty} \|\dot{\vb}(t) \|_{\mathrm{F}}^2 =0$ since $\dot{\vb}=\va {\vc}^{\T} $.  

\textbf{Step 2}: We show $\lim_{t \rightarrow \infty} \dot{E}(t)=0$. Without loss of generality, we suppose that $\lim_{t \rightarrow \infty} \| \vc(t)\|_2^2 =0$. The case of $\lim_{t \rightarrow \infty} \| \va(t)\|_2^2 =0$ is similar. That is $\norm{\va(0)}_2 > \norm{\vc(0)}_2$. By conservation law,  we have $\lim_{t \rightarrow \infty} \| \dot{ \va}(t) \|_2^2 =0.$

Considering the second derivative of $c_{k'}$, we have 
\begin{equation*}
    \ddot{c}_{k'} = \sum_{k} \left( \sum_{l} b_{kl} c_l b_{kk'} + a_k^2 c_{k'} \right).
\end{equation*}
Thus $\lim_{t \rightarrow \infty}\|\ddot{\vc}(t)\|_2^2=0$ since $\lim_{t \rightarrow \infty} \norm{\vc(t)}_2^2 = 0$. Note that  $\dot{E}(t) \ge 0$ and $\lim_{t \rightarrow \infty} E(t)=0$. Recall  that $\liminf_{t \rightarrow \infty} \dot{E}(t)=0$ and $\dot{E} $ is bounded. And also we have $\| \dot{\vc }\|_2^2 \le M$. We claim that $\lim_{t \rightarrow \infty} \| \dot{\vc}(t)\|_2^2=0$, which implies 
\begin{equation}
\label{eq::LimitDireEnergy}
    \lim_{t \rightarrow \infty} \dot{E}(t)=0.
\end{equation}

Since $\lim_{t \rightarrow \infty} \| \vc(t)\| _2 =0$ and $\lim_{t \rightarrow \infty} \|\ddot{\vc}(t) \| _2 =0$.  Using Taylor expansion, we have for some $\varphi \in [t,t+1]$
\begin{equation*}
    c_k(t+1)=c_k(t)+ \dot{c_k} (t)+\frac{1}{2} \ddot{c_k}(\varphi) , \forall k,
\end{equation*}
which implies $\lim_{t \rightarrow \infty} \|\dot{\vc} (t)\| _2 =0$. Therefore the assertion holds.

\textbf{Step 3}: We show that Equation \eqref{eq::LimitDireEnergy} contradicts with Assumption \ref{assump::BlowUpIniData}. By direct calculation, we have
\begin{align*}
    \frac{\D}{\D t} \| \dot{\va} \|_2^2 &=\frac{\D }{\D t} (\vc^{\T} \vb^{\T} \vb \vc) \\
    &=\dot{\vc}^{\T} \vb^{\T} \vb \vc + \vc^{\T} \dot{\vb}^{\T} \vb \vc +\vc^{\T} \vb^{\T} \dot{\vb} \vc+ \vc^{\T} \vb^{\T} \vb \dot{\vc}\\
    &= \va^{\T} \vb \vb^{\T} \vb \vc+ \vc^{\T} \vc \va^{\T} \vb \vc + \vc^{\T} \vb^{\T} \va \vc^{\T} \vc + \vc^{\T} \vb^{\T} \vb \vb^{\T} \va\\
    &= 2E  \| \vc \|_2^2 + 2 \dot{\va}^{\T} \vb \dot{\vc}
\end{align*}
and
\begin{align*}
    \frac{\D }{\D t} \| \dot{\vc} \|_2^2 & = \frac{\D}{\D t} (\va^{\T} \vb \vb^{\T} \va) \\
    &=\dot{\va}^{\T} \vb \vb^{\T} \va + \va^{\T} \dot{\vb} \vb^{\T} \vc + \va^{\T} \vb \dot{\vb}^{\T} \va+ \va^{\vT} \vb \vb^{\T} \dot{\va}\\
    &= \vc^{\T} \vb^{\T} \vb \vb^{\T} \va+ \va^{\vT} \va \vc^{\T} \vb^{\T} \va +\va^{\T} \vb \vc \va^{\T} \va + \va^{\T} \vb \vb^{\T} \vb \vc\\
    &=2E  \| \va \|_2^2 + 2 \dot{\va}^{\T} \vb \dot{\vc}.
\end{align*}
Thus
\begin{equation*}
    \frac{\D}{\D t} \| \dot{\va} \|_2^2 - \frac{\D}{\D t} \| \dot{\vc} \|_2^2 =2E (\|\vc(0)\| _2^2 -\|\va(0)\| _2^2).
\end{equation*}
Integrating both sides of the equation, we get
\begin{equation*}
    \lim_{t \rightarrow \infty} \| \dot{\va}(t) \|_2^2 - \| \dot{\vc}(t) \|_2^2 =\| \dot{\va}(0) \|_2^2 - \| \dot{\vc}(0) \|_2^2 -\|\vc(0)\|_2^2 ( \|\vc(0) \|_2^2 -\| \va(0) \|_2^2).
\end{equation*}
However, according to Assumption \ref{assump::BlowUpIniData} and the fact that $\lim_{t \rightarrow \infty} \|\dot{\vc}(t) \|_2 = 0$, we have that
\begin{equation*}
    \lim_{t \rightarrow \infty} \| \dot{\va}(t) \|_2^2  \neq 0.
\end{equation*}
Based on Equation (\ref{eq::DeriE}), we have 
\begin{equation*}
\begin{aligned}
        \lim_{t \rightarrow \infty} \dot{E}(t) &=\lim_{t \rightarrow \infty} \| \dot{\va}(t) \|_2^2 +\| \dot{\vc}(t) \|_2^2 +\| \va(t) \|_2^2 \|\vc(t) \|_2^2  \\
        &= \lim_{t \rightarrow \infty} \| \dot{\va}(t) \|_2^2  \neq 0.
\end{aligned}
\end{equation*}
It contradicts with Equation (\ref{eq::LimitDireEnergy}) which claims  $\lim_{t \rightarrow \infty} \dot{E} (t)=0$. This completes the proof.
\end{proof}

Based on Proposition \ref{prop::ConservationLaw}, the following corollary is trivial.

\begin{cor}
\label{coro::norm}
    Suppose that Assumption \ref{assump::BlowUpIniData} holds. Consider dynamical system \eqref{eq::SimpEffectiveModel}. The norm of  $\va$, $\vb$ and $\vc$ will diverge to infinity. That is 
    \begin{equation}
        \left\{
        \begin{aligned}
            \lim_{t \rightarrow T^*} \|\va \|_2= + \infty,\\
            \lim_{t \rightarrow T^*} \norm{\vb}_{\mathrm{F}} = + \infty,\\
            \lim_{t \rightarrow T^*} \|\vc \|_2= + \infty.
        \end{aligned}
        \right.
    \end{equation}
\end{cor}

\begin{proof}
Theorem \ref{thm::BlowUp} shows that the energy will diverge to infinity. If any of the norms $\norm{\va}_2$,$\norm{\vb}_{\mathrm{F}}$,$\norm{\vc}_2$ is bounded, then all the norms are bounded by conservation laws. This contradicts with the fact that the energy blows up. 
\end{proof}

\section{Final Stage Condition}\label{sec::FinalStaCond}
We have proved that energy will blow up under Assumption \ref{assump::BlowUpIniData}. It implies the  effective dynamics failed in finite time. The next question is how  the effective dynamics affects the emergency of condensation and whether there exist  observables to help us characterize condensation. 

Generally speaking, the condense phenomenon implies that the final stage of the dynamics will have good properties. Based on this observation, we propose  an assumption of final stage and verify its effectiveness using experimental and theoretical methods. In particular, we theoretically prove that the solution of the effective dynamics has specific properties, which is  to some extent a sufficiency argument. The necessity argument is quite difficult in theory. But experimental results  provide us  a strong implication that this condition maybe also necessary.

\begin{assump}[final stage condition]
    \label{assump::FinalStageCond}
    The parameters satisfy the \textbf{final stage condition} at time $t$. That is

    \begin{enumerate}
        \item  For every index $i \in [m]$ , we have $c_i \va^{\T} \vb^i >0$ and $a_i \vb_i \vc>0$.
        \item  For every index $i,j \in [m]$, we have $ \langle c_i \vb^i,c_j \vb^j \rangle > 0 $ and $ \langle a_i \vb_i, a_j \vb_j \rangle > 0$.
    \end{enumerate}
\end{assump}

First, we observe the role which final stage condition played in training dynamics by experiments.
\subsection{Experimental Results}\label{sec::ExperResul}
We present a numerical example using the following target function
\begin{equation*}
    f(x) = \tanh (x).
\end{equation*}
The training data are uniformly sampled from $[-\pi ,\pi]$ with sample size $100$. We use a $1-200-200-1$ neural network  with tanh as activation function. Parameters are initialized from $\fN (0,\frac{1}{200^4})$.

\begin{figure}[h]
    \centering
    \begin{subfigure}{0.24\textwidth}
        \centering
        \includegraphics[width=\textwidth]{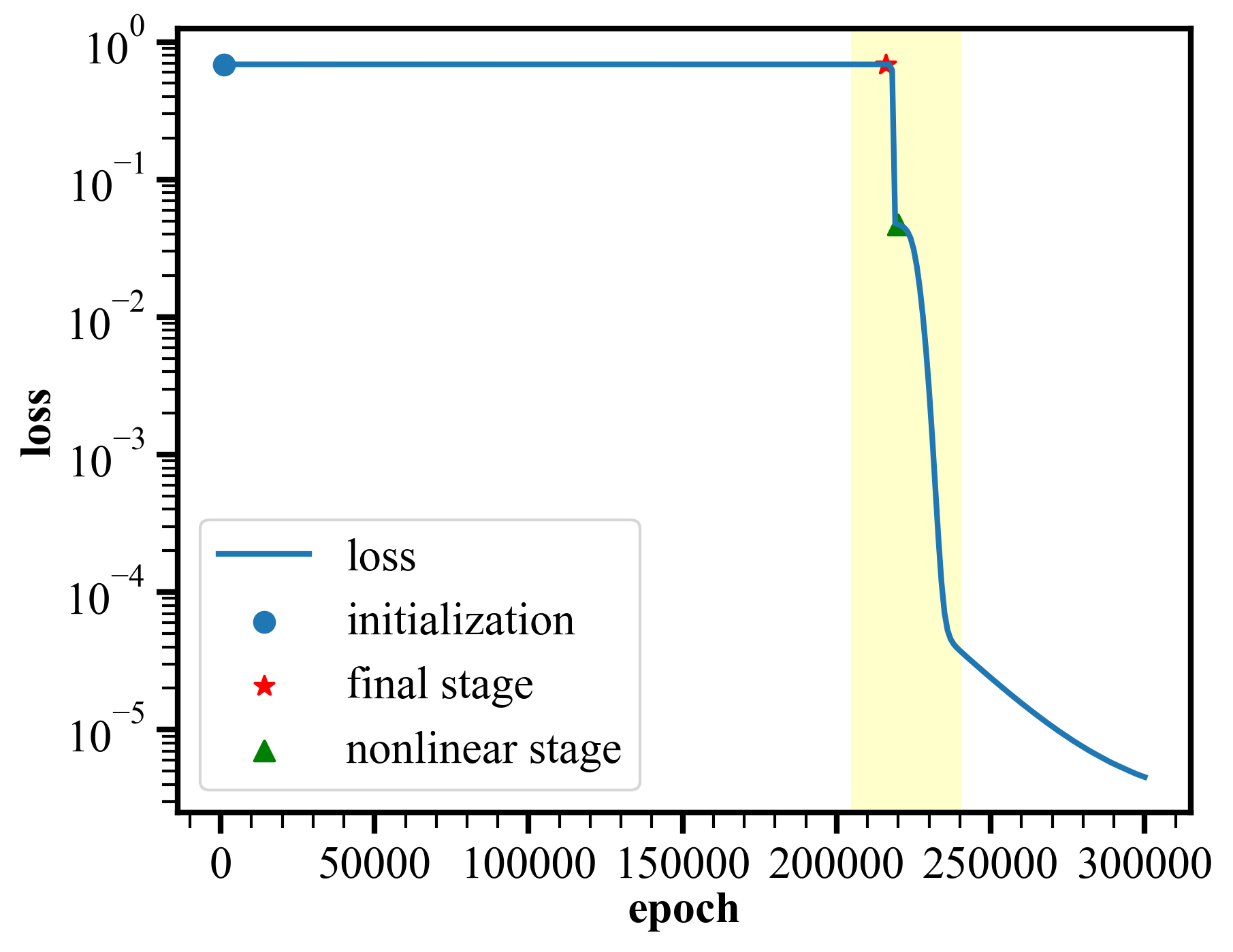}
        \caption{loss}
        \label{fig::Loss}
    \end{subfigure}
    \begin{subfigure}{0.24\textwidth}
        \centering
        \includegraphics[width=\textwidth]{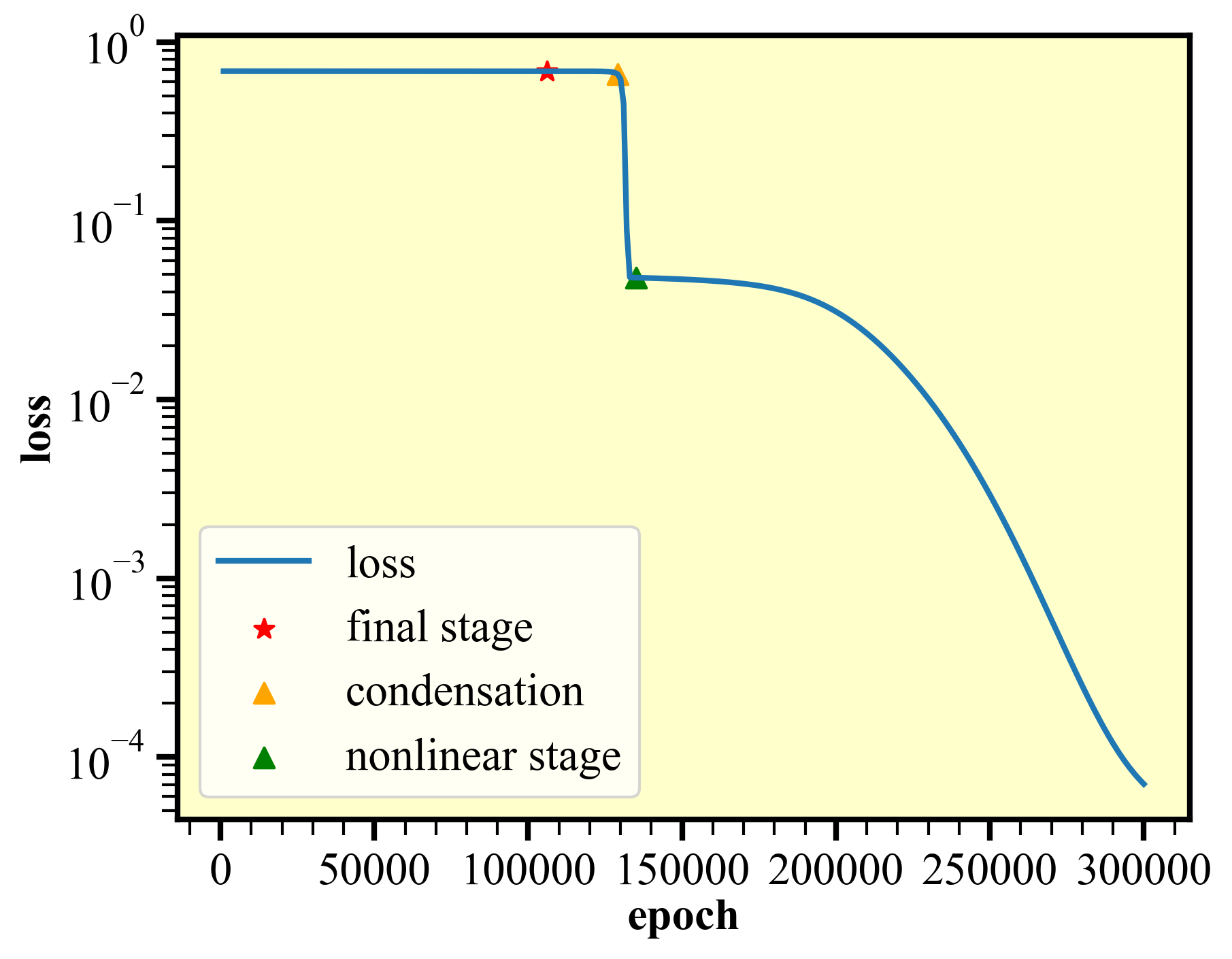}
        \caption{final state}
        \label{fig::LossSmallLr}
    \end{subfigure}
        \begin{subfigure}{0.23\textwidth}
        \centering
        \includegraphics[width=\textwidth]{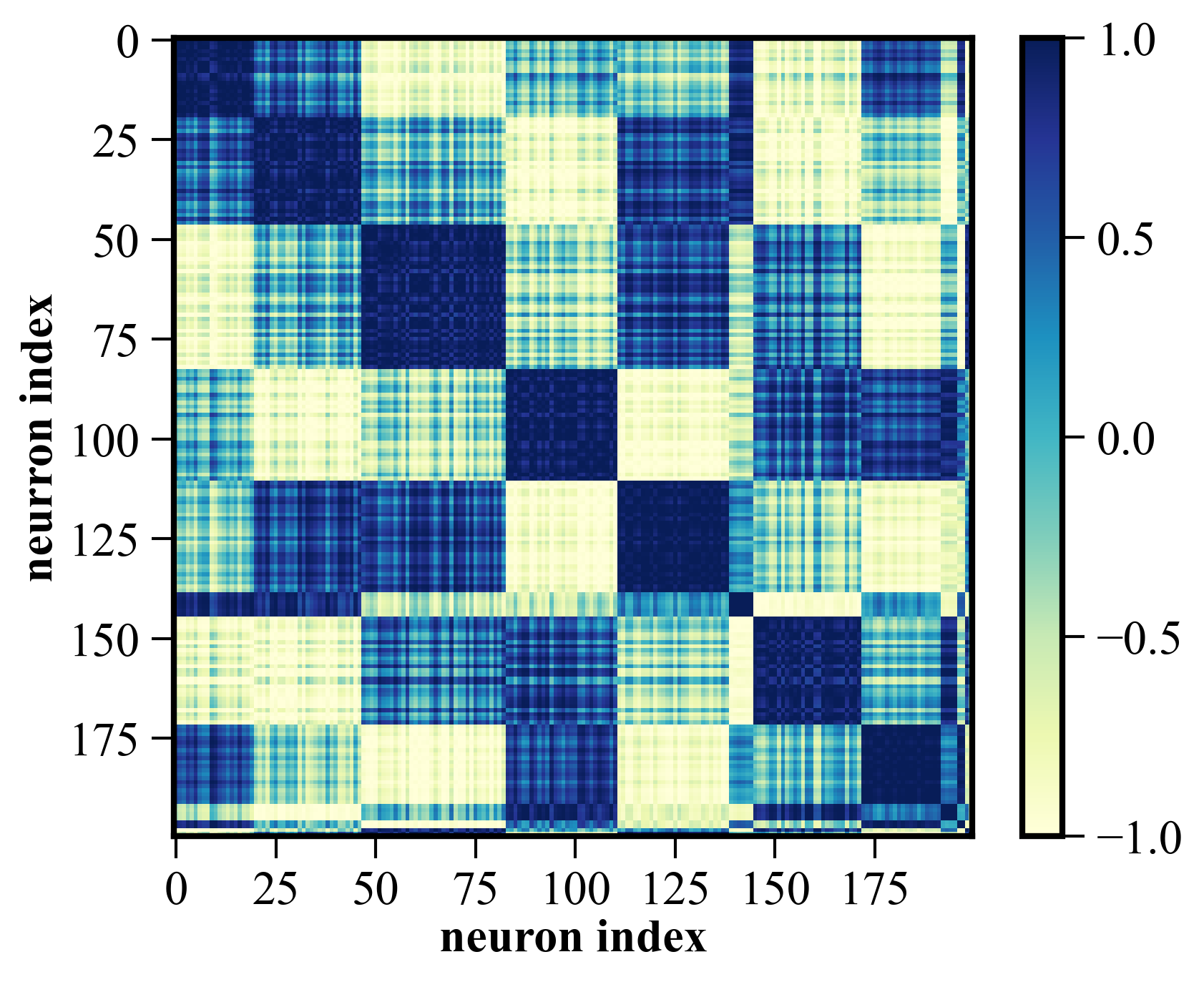}
        \caption{initialization}
        \label{fig::Initial}
    \end{subfigure}
    \begin{subfigure}{0.23\textwidth}
        \centering
        \includegraphics[width=\textwidth]{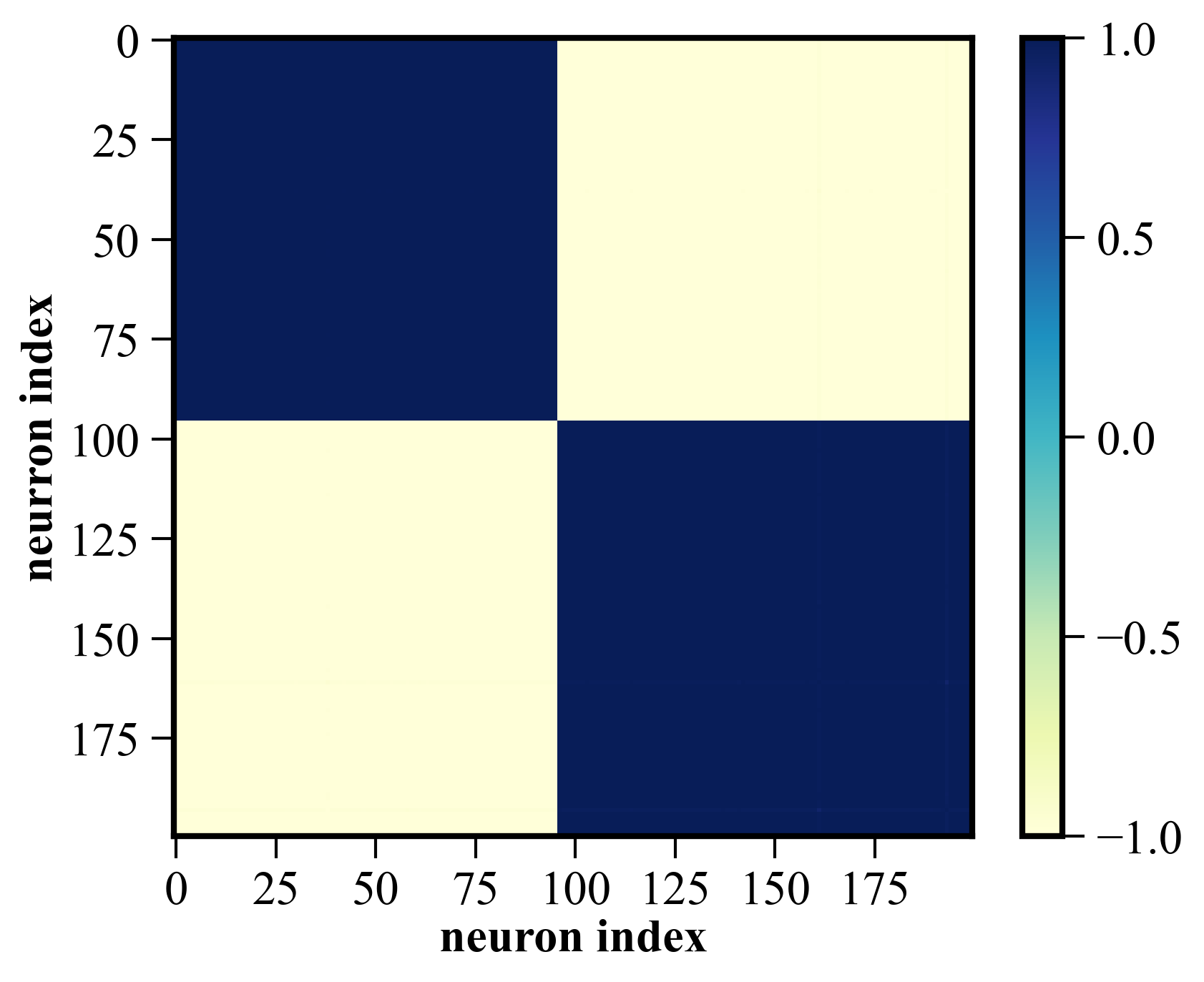}
        \caption{condensation}
        \label{fig::Condensation}
    \end{subfigure}
    \caption{Loss curve and condensation phenomenon via gradient descent. The left two plots display the loss changes for the initially trained NN with a learning rate of $5 \times 10^{-3}$ and the retrained NN loaded with the model from iteration $205,000$ using a learning rate of $5 \times 10^{-4}$. The shaded region indicates where effective model fails. 
    The two plots on the right show the cosine similarity of the innermost layer parameters  $\vW^{[1]}$ during initialization and condensation respectively. }
    \label{fig::ExperiCondensation}
\end{figure}
 Figure \ref{fig::ExperiCondensation}  presents empirical demonstrations of our conclusions from Theorem \ref{thm::Condense}. It shows that innermost layer parameters $\vW^{[1]}$ converge to target direction $\bm{v}$. Moreover, it implies the potential of the final stage condition: this is almost a critical indicator of the collapse of effective dynamics. Informally speaking, the prolonged plateau in the loss curve is, in fact, the process that effective dynamics searches  the final stage condition.
\begin{figure}[h]
    \centering
    \begin{subfigure}{0.3\textwidth}
        \centering
        \includegraphics[width=\textwidth]{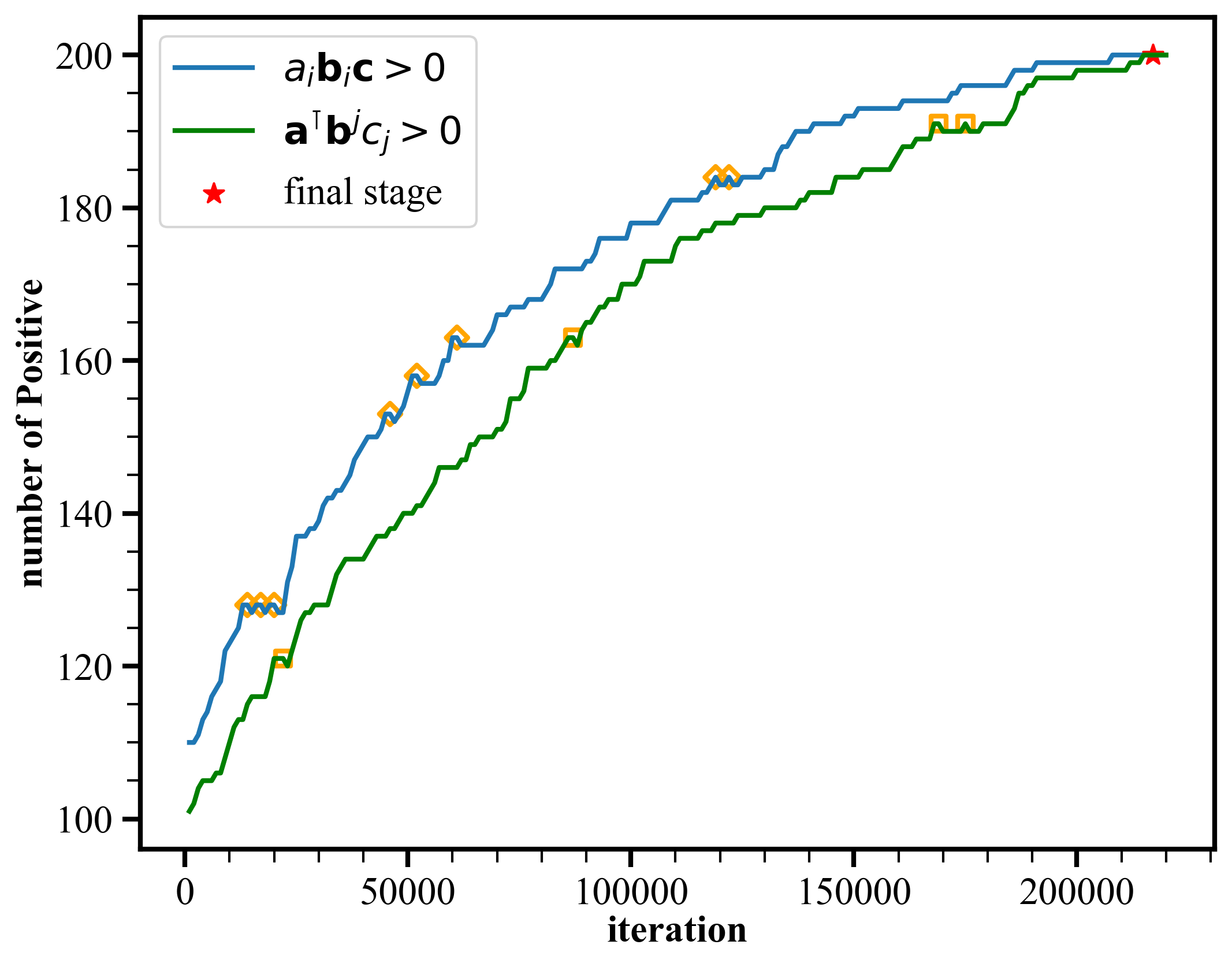}
        \caption{positive number changes}
        \label{fig::PositiveNumChange}
    \end{subfigure}
    \begin{subfigure}{0.3\textwidth}
        \centering
        \includegraphics[width=\textwidth]{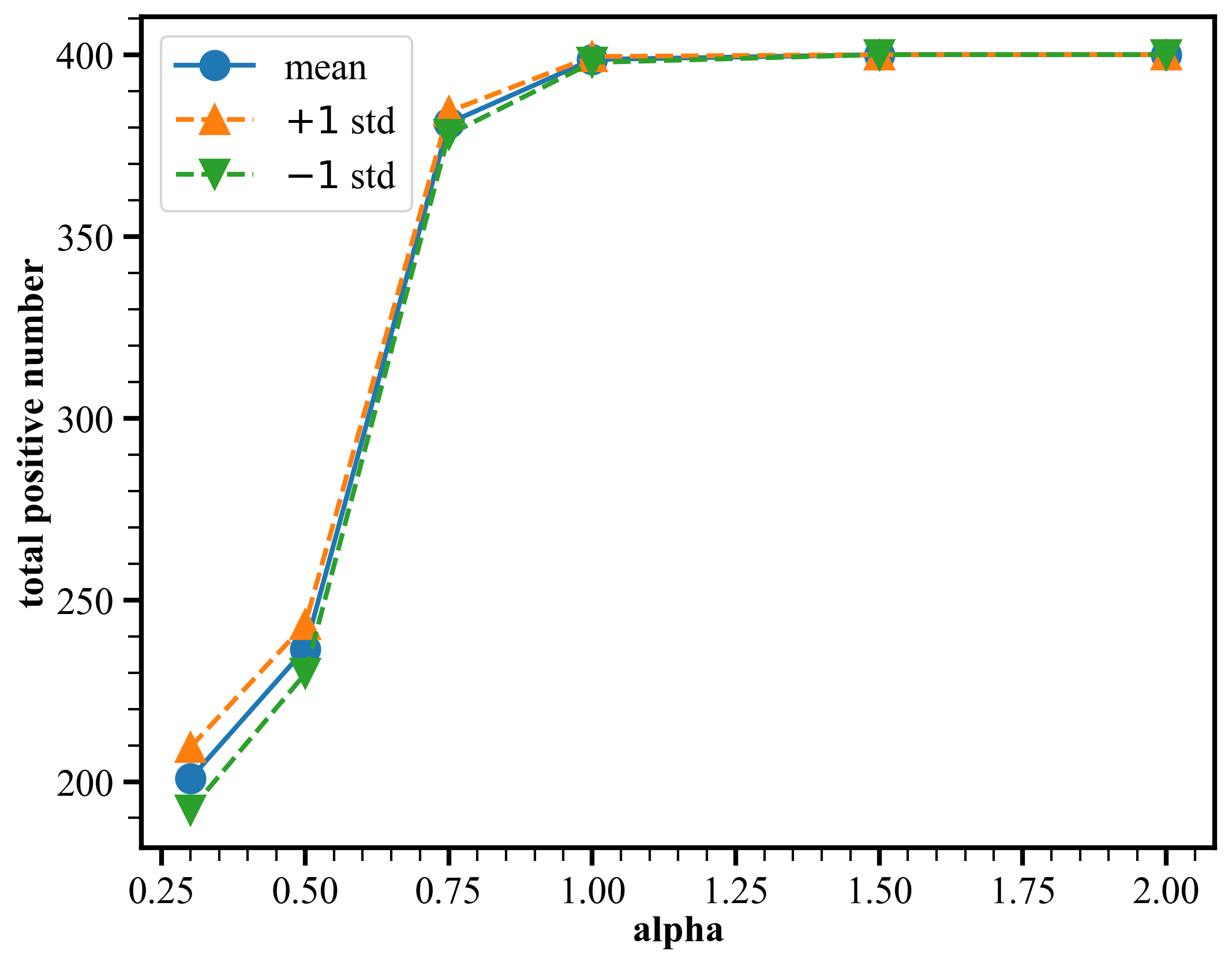}
        \caption{smaller initialization}
        \label{fig::SmallerIni}
    \end{subfigure}
    \hfill
    \caption{Variations of first condition of Assumption \ref{assump::FinalStageCond} respect to training process and smaller initialization. The left plot shows the changes of first condition under initialization $\fN (0,\frac{1}{200^4})$. The right plot shows the max sum of positive numbers during whole training process under different initialization $\fN (0,\frac{1}{200^{2 \alpha}})$which is parametrized by $\alpha$.  }
    \label{fig::ExperimentFSC}
\end{figure}

 To further illustrate the above points, we proceed with the following experiment. Figure \ref{fig::PositiveNumChange} illustrates the variations of statement 1 of Assumption \ref{assump::FinalStageCond}. It implies that, despite the overall increasing trend in the indicators proposed by Assumption \ref{assump::FinalStageCond}, there are also instances of peaks or declines. This  indicates difficulties in theoretical analyses on this platform. Figure \ref{fig::SmallerIni} shows that smaller initialization  leads to a more stable final stage condition, providing evidence for the validity of the Assumption \ref{assump::FinalStageCond} under small initialization.

Similar discussion applies to the scene of matrix completion. To illustrate this viewpoint, we model gradient descent of matrix factorization and introduce final stage condition for matrix completion(details in Appendix \ref{sec::FinalStageMC}). Generally speaking, in matrix completion, we will consider the gradient ascent of energy $E$ defined by 
\begin{equation}
    E = \sum_{i=1}^n \lambda_i \va_i^{\T} \vb \vc_{i}.
\end{equation}
Then the final stage condition will be generalized in following sense.
\begin{assump}[final stage condition for matrix completion]
    The parameters satisfy the final stage condition at time $t$. That is 
    \begin{enumerate}
        \item For all index $i \in [n]$ , $j \in [m]$ , we have $ \lambda_i c_{i,j} \va_i^{\T} \vb^j >0$ and $ \lambda_i a_{i,j} \vb_j \vc_i >0$.
        \item For all index $i \in [n] $ , $j,k \in [m]$, we have $ \langle c_{i,k} \vb^k, c_{i,j} \vb^j \rangle > 0 $ and $ \langle a_{i,k} \vb_k, a_{i,j} \vb_j \rangle > 0$.
        \item For all index $i,j \in [n]$ , $k,l \in [m]$, we have $ \lambda_i \lambda_j a_{i,k} a_{j,k} c_{i,l} c_{j,l} > 0$. 
    \end{enumerate}
\end{assump}

\begin{figure}[h]
    \centering
    \begin{subfigure}{0.35\textwidth}
        \centering
        \includegraphics[width=\textwidth]{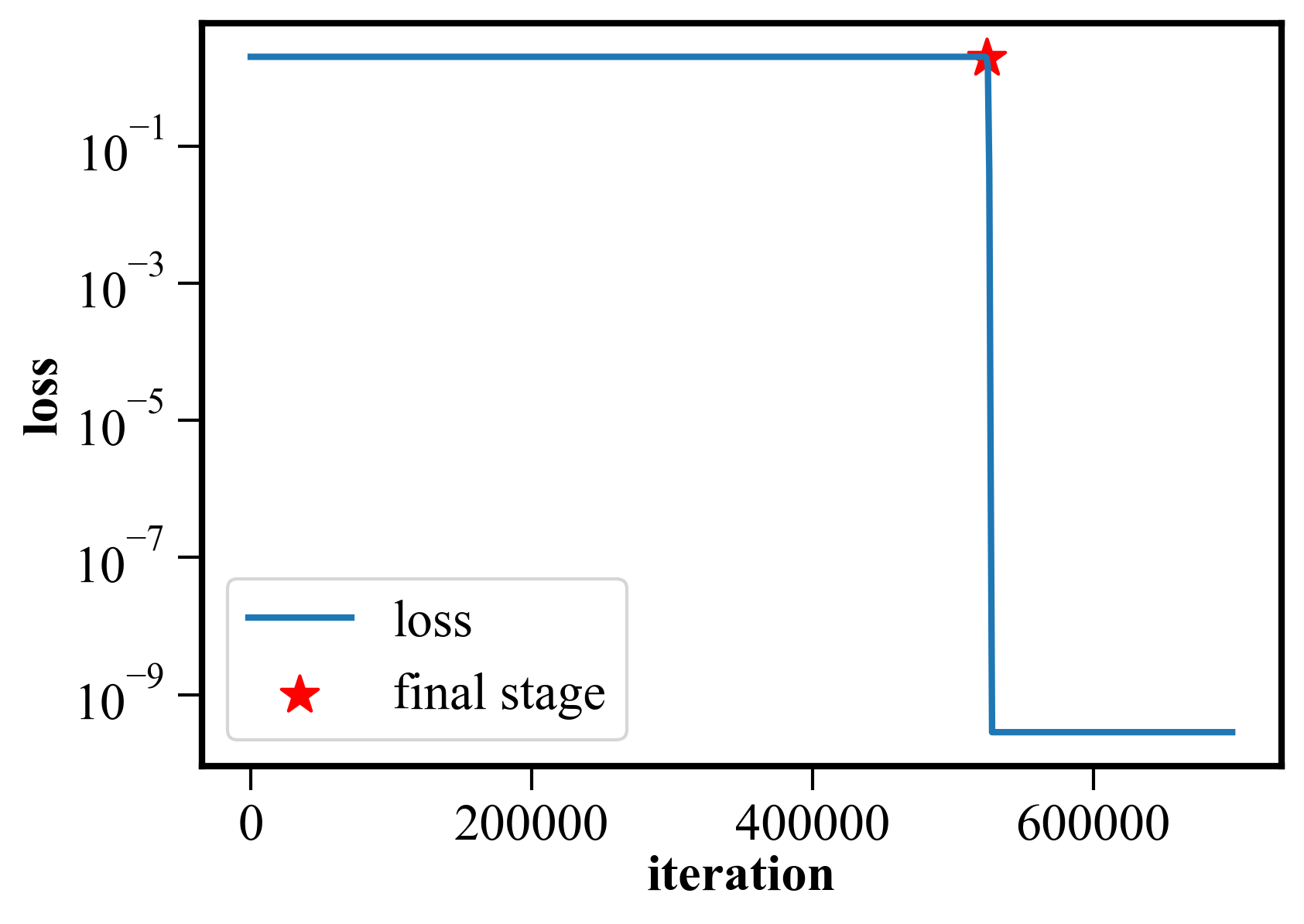}
        \caption{loss}
        \label{fig::LossMC}
    \end{subfigure}
    \begin{subfigure}{0.35\textwidth}
        \centering
        \includegraphics[width=\textwidth]{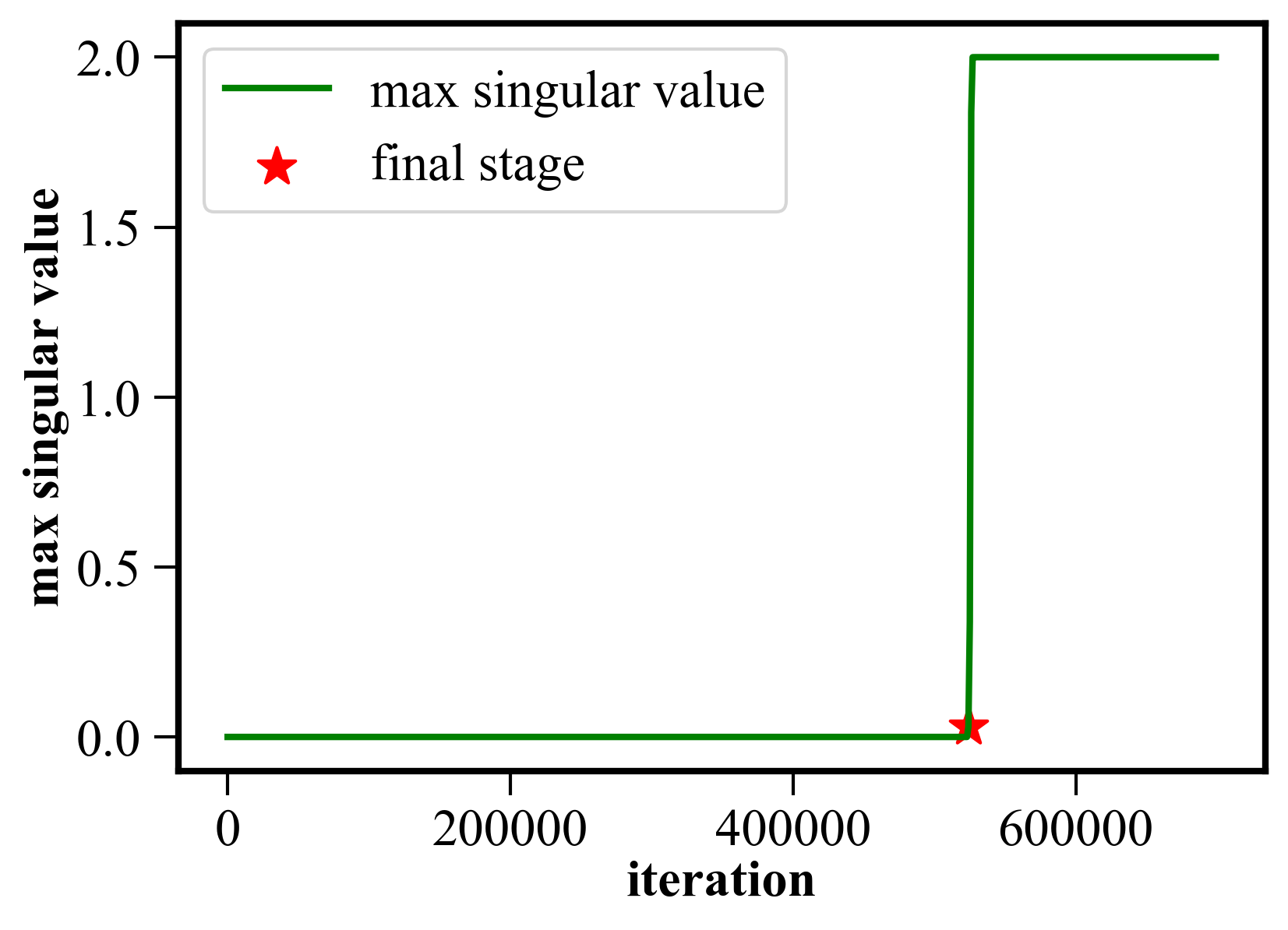}
        \caption{singular value}
        \label{fig::SinValMC}
    \end{subfigure}
    \hfill
    \caption{Loss and max singular value.  The left and right figures  show the changes of loss and the maximum singular value respectively.  }
    \label{fig::ExperimentFSCMC}
\end{figure}

We takes following numerical example to illustrate the existence of final stage condition in matrix factorization. Without loss of generality, the matrix we take to be recovered is a diagonal matrix with only two non-zero elements 1 and -1 on its diagonal. That is 
\begin{equation*}
\left(
\begin{array}{cccc}
    1 & 0  & \cdots & 0\\
    0 & -1 && \\
    \cdots & & \cdots & \\
    &&& 0 \\
\end{array}
\right).
\end{equation*}
The NN we use is a linear network $\vW = \vW^{[3]} \vW^{[2]} \vW^{[1]} $, where $\vW^{[3]},\vW^{[2]},\vW^{[1]} \in {\sR}^{100 \times 100} $. The parameters are initialized from $\fN (0,\frac{1}{100^6})$. Figure \ref{fig::ExperimentFSCMC} shows that similar phenomenon exists in matrix completion. This indicates the generality of the final stage condition.

\subsection{Theoretical Results}\label{sec::TherRes}
In this section, we prove the main theorems which characterize the condensation.  In retrospect of  the proof of Theorem \ref{thm::BlowUp},  Equation  \eqref{eq::LowBd} provides a  lower bound that leads to the presence of explosion. However, this inequality leaves the precise growth rate of energy $E$ undetermined. 
The key idea here is that Assumption \ref{assump::FinalStageCond} can give us an upper limit on how fast the energy can grow. Once we understand this growth rate, we can then move forward with proving the main theorems.

 Let us  define condensation in the effective dynamics based on our derivation of the effective dynamics.  The vector $\vc$ is identified as the projection of the innermost layer $\vW^{[1]}$ onto the target vector $\bm{v}$. This leads to the formal definition of condensation.

\begin{defi}[condensed solution]
\label{defi::CondSolu}
The dynamical system  \eqref{eq::SimpEffectiveModel} is said to possess a \textbf{condensed solution} if, for every index $i \in [m]$, the following limit holds:
\begin{equation}
\lim_{{t \to T^*}} c_i = \pm \infty.
\end{equation}
\end{defi}

Then we begin our proof by the following proposition which implies the definition of the final stage is reasonable.

\begin{prop}[induction]
\label{prop::Induction}
Consider dynamical system \eqref{eq::SimpEffectiveModel}. If Assumption \ref{assump::FinalStageCond} holds at some time $t_0 $ with $t_0 < T^*$, then Assumption \ref{assump::FinalStageCond}  will hold at $t_0 < t <T^*$. 
\end{prop}

\begin{proof}
    First, we consider the second condition in Assumption \ref{assump::FinalStageCond}. By direct calculation, we have   
\begin{equation*}
\label{eq::CrossTerm1}
    \frac{\D}{\D t} \left\langle c_i \vb^i, c_j \vb^j \right\rangle 
    =\left(c_j \va^{\T} \vb^j \right) \left(c_i^2 + \frac{1}{c_j^2} c_i c_j {\vb^i}^{\T} \vb^j \right) 
    +\left( c_i \va^{\T} \vb^i \right) \left(c_j^2 + \frac{1}{c_i^2}c_i c_j {\vb^i}^{\T} \vb^j \right)
\end{equation*}
and 
\begin{equation*}
\label{eq::CrossTerm2}
     \frac{\D}{\D t} \langle a_i \vb_i, a_j \vb_j \rangle 
     =\left(a_j \vb_j \vc \right) \left(a_i^2 + \frac{1}{a_j^2} a_i a_j \vb_i \vb_j^{\T} \right) 
     +\left(a_i \vb_i \vc \right) \left(a_j^2 + \frac{1}{a_i^2}a_i a_j \vb_i \vb_j^{\T} \right).
\end{equation*}
By Assumption \ref{assump::FinalStageCond}, we know the above equations are larger than $0$. So $ \langle c_i \vb^i,c_j \vb^j \rangle $ and $ \langle a_i \vb_i, a_j \vb_j \rangle $ will be monotonically increasing since $t_0$.

Calculating the derivative of left hand side of first condition, we have
\begin{equation*}
    \frac{\D}{\D t} c_i \va^{\T} \vb^i= \dot{c_i}^2 + \sum_{j=1}^m c_i c_j {\vb^j}^{\T} \vb^i +c_i^2 \| \va \|_2^2
\end{equation*}
and
\begin{equation*}
    \frac{\D}{\D t} a_i \vb_i \vc =\dot{a_i}^2 + \sum_{j=1}^m a_i a_j \vb_i \vb_j^{\T} +a_i^2 \| \vc \|_2^2.
\end{equation*}
 Hence, $c_i \va^{\T} \vb^i$ and $a_i \vb_i \vc$ will also increase monotonically since $t_0$. Therefore the final stage condition will hold until $T^*$.
\end{proof}

Next, we analyze the angle relation between $\va$ and its derivative $\dot{\va}$. For the simplicity of proof and description, we adopt a standardized notation to represent angles between distinct vectors throughout the ensuing discussion.
\begin{defi}
\label{defi::Angles}
    Let $\xi_{ij}(t)$ denote the angle between the vectors $c_i (t) \vb^i (t)$ and $c_j (t) \vb^j (t)$, and $\psi_i (t)$ denote the angle between the vectors $\dot{\va} (t)$ and $c_i (t) \vb^i (t)$. Let $\varphi_i (t)$ denote the angle between $\va (t)$ and $c_i (t) \vb^i (t)$, while $\zeta (t)$ denote the angle between $\va(t)$ and $\dot{\va} (t)$.  In subsequent expressions, the variable $t$ will be omitted unless there is a specific emphasis on the temporal change of angles.
\end{defi}

We divide the entries of vector $\vc$ into two classes according to whether  their limit is finite. 

\begin{prop}
\label{prop::Angle}
Suppose that Assumption \ref{assump::FinalStageCond} holds. Consider the effective dynamics \eqref{eq::SimpEffectiveModel}. The indices $[m]$ can be partitioned into two disjoint classes, denoted by $C_1= \{ i_1 , \dots , i_k \} \neq \varnothing$ and $C_2=[m] \setminus C_1$.  The partition satisfies the following properties: 

\begin{enumerate}[label=(\roman*)]
\item For each $i \in [m]$, the limits of $c_i$ exist. In particular, 
    \begin{equation}
    \lim_{t \rightarrow T^*} c_i^2=
    \left\{
    \begin{aligned}
    &+ \infty, \quad &i \in C_1,\\
    &c_i^{*}, \quad &i \in C_2.\\
    \end{aligned}
   \right.
    \end{equation}
\item The angle $\xi_{ij}$  between the vectors $c_i\vb^i$ and $c_j \vb^j$, as defined in Definition \ref{defi::Angles}, fulfills the condition:
\begin{equation}
\label{eq::Cos1}
\lim_{t \rightarrow T^*} \cos \xi_{ij} = 1, \quad \text{for } i, j \in C_1.
\end{equation}
\item The following limits exist
\begin{equation}
    \label{eq::AdotVsCnorm}
        \lim_{t \rightarrow T^*} \frac{ \norm{\dot{\va}}_2}{ \norm{\vc}_2^2 } =  \lim_{t \rightarrow T^*} \frac{ \norm{\dot{\va}}_2}{ \norm{\va}_2^2 }  = 1 .
    \end{equation}
\end{enumerate}
\end{prop}

\begin{proof}
1.
    First, we find that for every index $i$ the $c_i^2$ is monotonically increasing. So their limits exist. We define the index set of the parameters that tend to infinity as $C_1$ and the others as $C_2$. Based on Corollary \ref{coro::norm}, we know that $C_1 \neq \varnothing $. Property 1 is automatically satisfied due to our partition.

2. We introduce new variables
    \begin{equation*}
    \left \{
    \begin{aligned}
    & p= \langle c_i \vb^i , c_j \vb^j \rangle,\\
    & q=c_i^2 c_j^2.
    \end{aligned}
    \right.
    \end{equation*}
    
    According to the proof of  Proposition \ref{prop::Induction}, we find that
    \begin{equation}
    \label{eq::DyDt}
    \begin{aligned}
    \frac{\D p }{\D t}  
    &=\left(c_j \va^{\T} \vb^j \right) \left(c_i^2 + \frac{1}{c_j^2} c_i c_j {\vb^i}^{\T} \vb^j \right) 
    +\left( c_i \va^{\T} \vb^i \right) \left(c_j^2 + \frac{1}{c_i^2}c_i c_j {\vb^i}^{\T} \vb^j \right) \\
    & = \left(c_j \va^{\T} \vb^j c_i^2 + c_i \va^{\T} \vb^i c_j^2 \right) \left(1 + \frac{ c_i c_j {\vb^i}^{\T} \vb^j}{c_i^2 c_j^2 } \right) =  \left(c_j \va^{\T} \vb^j c_i^2 + c_i \va^{\T} \vb^i c_j^2 \right) \left(1 + \frac{p}{q} \right).
    \end{aligned}
    \end{equation}

Thanks to \eqref{eq::SimpEffectiveModel}, we obtain
    \begin{equation}
    \label{eq::DxDt}
        \frac{\D q}{\D t} = 2 c_i \va^{\T} \vb^i  c_j^2 + 2 c_j \va^{\T} \vb^j c_i^2 .
    \end{equation}
Combining Equation \eqref{eq::DyDt} and  Equation \eqref{eq::DxDt}, we obtain
    \begin{equation}
    \label{eq::DyDx}
        \frac{\D p}{\D q}=\frac{1}{2} \frac{p}{q}+\frac{1}{2}.
    \end{equation}

    Let $u = p/q $. Note that $\frac{\D p}{\D q }  = \frac{\D }{\D q} (uq) = q \frac{\D u}{\D q} + u $.
    Combining this with the right hand of Equation \eqref{eq::DyDx}, we get
    \begin{equation}
    \label{eq::DxDu}
         \frac{\D q}{q}= \frac{2 \D u }{1-u}.
    \end{equation}

    The Equation (\ref{eq::DxDu}) can be solved explicitly. 
    \begin{equation}
        \ln |q(t)| - \ln|q(t_0)| = -2 \ln|u(t)-1| + 2 \ln|u(t_0) -1|.
    \end{equation}
    For $i,j \in C_1$, we have $\lim_{t \rightarrow T^* } u(t) =1 $  since $q $ tends to infinite as $t$ tends to $T^*$.
    
    By definition, 
    \begin{equation}
    \label{u_defi}
        u= \frac{p}{q} = \frac{ \langle c_i \vb^i , c_j \vb^j \rangle }{c_i^2 c_j^2 }= \frac{ \norm{\vb^i}_2 \norm{\vb^j}_2 }{|c_i| | c_j|} \cos \xi_{ij}.
    \end{equation}
    Using the conservation laws, we have 
    \begin{equation*}
        \norm{\vb^i}_2^2 (t) - \norm{\vb^i}_2^2 (0) = c_i^2(t) -c_i^2(0).
    \end{equation*}
    For $i \in C_1$, we have 
    \begin{equation}
    \label{bi_ci}
        \lim_{t \rightarrow T^*} \frac{\norm{\vb^i}_2^2}{c_i^2} =1.
    \end{equation}
    Combining Equation (\ref{u_defi}) and (\ref{bi_ci}), we get  
    \begin{equation}
        \lim_{t \rightarrow T^*}  \cos \xi_{ij}=1, \ i,j \in C_1.
    \end{equation}
    This finishes the proof of statement (\romannumeral 2).

    3.
    Finally, we calculate the norm $ \norm{\dot{\va}}_2^2 $. By definition, we obtain
    \begin{equation*}
         \langle \dot{\va} , \dot{\va} \rangle =\sum_{i=1}^m \sum_{j=1}^m \langle   c_i \vb^i ,  c_j \vb^j \rangle.
    \end{equation*}
    We divide the sum into three parts due to the boundedness of  entries of  $\vc$.
    \begin{equation*}
        \begin{aligned}
             \langle \dot{\va}, \dot{\va} \rangle &=\sum_{i \in C_1} \sum_{j \in C_1} c_i^2 c_j^2 \frac{\norm{\vb^i}_2}{|c_i|} \frac{\norm{\vb^j}_2}{|c_j|} \cos \xi_{ij} + 2 \sum_{i \in C_1 } \sum_{j \in C_2} c_i  c_j  \norm{\vb^i}_2 \norm{\vb^j}_2 \cos \xi_{ij}\\
            &+  \sum_{i \in C_2 } \sum_{j \in C_2} c_i  c_j \norm{\vb^i}_2 \norm{\vb^j}_2 \cos \xi_{ij}\\
            &=\sum_{i \in C_1} \sum_{j \in C_1} c_i^2 c_j^2 + \sum_{i \in C_1} \sum_{j \in C_1} c_i^2 c_j^2 \left( \frac{\norm{\vb^i}_2}{|c_i|} \frac{\norm{\vb^j}_2}{|c_j|} \cos \xi_{ij} -1 \right)\\
            &+ 2 \sum_{i \in C_1 } \sum_{j \in C_2} c_i  c_j \norm{\vb^i}_2 \norm{\vb^j}_2 \cos \xi_{ij} + \sum_{i \in C_2 } \sum_{j \in C_2} c_i  c_j \norm{ \vb^i }_2 \norm{\vb^j}_2 \cos \xi_{ij}.
        \end{aligned}
    \end{equation*}
    Since $\lim_{t \rightarrow T^*} \frac{\norm{\vb^i}_2}{|c_i|} \frac{\norm{\vb^j}_2}{|c_j|} \cos \xi_{ij}=1 $, we have 
    \begin{equation}
    \label{a_dot1}
     \lim_{t \rightarrow T^*} \frac{ \langle \dot{\va} , \dot{\va} \rangle}{ \left(  \sum_{i \in C_1} c_i^2 \right)^2} = 1.   
    \end{equation} 
    Based on statement (\romannumeral 1), we obtain 
    \begin{equation}
    \label{a_dot2}
        \lim_{t \rightarrow T^*} \frac{\sum_{i \in C_1} c_i^2}{\norm{\vc}_2^2}=1.
    \end{equation}
    Combining Equation (\ref{a_dot1}), (\ref{a_dot2}) and conservation law, we have 
    \begin{equation}
        \lim_{t \rightarrow T^*} \frac{ \norm{\dot{\va}}_2}{ \norm{\vc}_2^2 } = \lim_{t \rightarrow T^*} \frac{ \norm{\dot{\va}}_2}{ \norm{\va}_2^2 } =1 .
    \end{equation}
    This finishes the proof of statement (\romannumeral 3).
\end{proof}

Proposition \ref{prop::Angle} describes the angle between $c_i \vb^i$ and $c_j \vb^j$. Since $\dot{\va}$ is a linear combination of $c_i \vb^i$, we immediately have following corollary.

\begin{cor}
\label{coro::Dota_ci_bi}
Suppose that Assumption \ref{assump::FinalStageCond} holds. Consider the effective dynamics \eqref{eq::SimpEffectiveModel} and recall index class defined in Proposition \ref{prop::Angle}. The angle $\psi_i$ between the vectors $\dot{\va}$ and $c_i \vb^i$, as defined in Definition \ref{defi::Angles}, satisfies:
    \begin{equation}
        \lim_{ t \rightarrow T^*} \cos \psi_i  =1, \quad i \in C_1.
    \end{equation}  
\end{cor}

\begin{proof}
By definition,
\begin{equation*}
    \cos \psi_i = \frac{\langle c_i \vb^i, \sum_{j=1}^m c_j \vb^j \rangle}{ \norm{c_i \vb^i }_2  \norm{\dot{\va} }_2 }.
\end{equation*}
Recall the definition of $\xi_{ij}$, the above equation can be 
reformulated as follows:
\begin{equation*}
\begin{aligned}
     \cos \psi_i &  =\frac{\sum_{j=1}^m |c_i| |c_j| \norm{\vb^i}_2 \norm{\vb^j}_2 \cos \xi_{ij}}{\| c_i \vb^i \|_2 \|\dot{\va} \|_2} \\
            &= \frac{\sum_{j=1}^m |c_j| \norm{\vb^j}_2 \cos \xi_{ij}}{\norm{\dot{\va}}_2} \\
            &= \frac{\sum_{j \in C_1} c_j^2 \frac{\norm{\vb^j}_2}{|c_j|} \cos \xi_{ij} + \sum_{j \in C_2} |c_j| \norm{\vb^j}_2 \cos \xi_{ij}}{\norm{\dot{\va}}_2 }.
\end{aligned}
\end{equation*}
   According to  Equation (\ref{eq::Cos1}) and (\ref{eq::AdotVsCnorm}), we have  $\lim_{ t \rightarrow T^*} \cos \psi_i =1$. This completes the proof.
\end{proof}  

So far, we have characterized some properties of $c_i \vb^i$ which is component of $\dot{\va}$. We have shown that some of them will have the same direction when $t$ tends to $T^*$. However, it is not enough for our seek for a upper bound for energy $E$. Luckily,  based on Corollary \ref{coro::Dota_ci_bi}, we can analyze the angle between $\va$ and $c_i \vb^i$ which provides an upper bound. Before this, we give the following proposition. The subsequent proposition demonstrates an extension of statement 1 of Assumption \ref{assump::FinalStageCond}, going beyond the condition of $c_i \mathbf{a}^{\T} \mathbf{b}^i$ being greater than zero to include additional angle-related information.

\begin{prop}
\label{prop::4}
Suppose that Assumption \ref{assump::FinalStageCond} holds. Consider the effective dynamics \eqref{eq::SimpEffectiveModel} and recall index class defined in Proposition \ref{prop::Angle}.
There exists constants $T_1 \in (t_0,T^*)$ and $\Theta_1 \in [0,\frac{\pi}{2}) $ such that for each index $i \in C_1 $, the follow inequality holds:
    \begin{equation*}
        \cos \varphi_i \ge \cos \Theta_1,\quad t \in (T_1 ,T^*).
    \end{equation*}
\end{prop}
\begin{proof}
It is sufficient to prove the statement for any fixed $i \in C_1$ due to the finiteness of $|C_1|$.
In Proposition \ref{prop::Induction}, we have shown that $\langle c_i \vb^i, \va \rangle > 0 $ for $t \in (t_0,T^*)$, which implies 
    \begin{equation}
        \cos \varphi_i >0, \quad t \in (t_0,T^*).
    \end{equation}
    Hence we can focus on its square, i.e., $\cos^2 \varphi_i 
        =\frac{\langle \va, \vb^i \rangle^2}{\| \va \|_2^2 \| \vb^i \|_2^2}$.
    By direct calculation, the derivation of $\cos^2 \varphi_i $ is 
    \begin{equation*}
    \frac{2 \langle \va,\vb^i \rangle ( \langle \dot{\va} , \vb^i \rangle + \langle \va, \dot{ \vb}^i \rangle) \| \va \|_2^2 \| \vb^i \|_2^2 - 2 \langle \va,\vb^i \rangle^2 (\langle \dot{\va} ,\va \rangle \| \vb^i \|_2^2 + \| \va \|_2^2 \langle \dot{\vb}^i ,\vb^i \rangle)}{(\|\va\|_2^2 \|\vb^i \|_2^2)^2}.
    \end{equation*}
    We can rewrite the numerator as 
    \begin{equation*}
         2 \norm{\vb^i}_2^2  \langle \va, \vb^i \rangle \left[ \langle \dot{\va} , \vb^i \rangle \| \va \|_2^2 - \langle \va, \vb^i \rangle \langle \dot{\va}, \va \rangle \right]  
        + 2 \norm{\va}_2^2 \langle \va , c_i \vb^i \rangle \left[ \| \va\|_2^2 \| \vb^i \|_2^2- \langle \va,\vb^i \rangle^2 \right].
    \end{equation*}
    The second term of above expression is obviously greater than zero by inequality of arithmetic and geometric means. Also we can rewrite the first term as 
    \begin{equation*}
        \frac{2 \|\vb^i\|_2^2}{c_i^2} \langle \va,c_i \vb^i \rangle \left[ \langle \dot{\va} ,c_i \vb^i \rangle \|\va \|_2^2- \langle \va, c_i \vb^i \rangle \langle \dot{\va}, \va \rangle \right].
    \end{equation*}
    We find that the first two factors $ \frac{2\|\vb^i\|_2^2}{c_i^2} \langle \va,c_i \vb^i \rangle$ of above expression  are positive. According to Definition \ref{defi::Angles}, the difference term can be reformulated as 
    \begin{equation*}
        \langle \dot{\va} ,c_i \vb^i \rangle \|\va \|^2- \langle \va, c_i \vb^i \rangle \langle \dot{\va}, \va \rangle=  \| \va \|_2^2 \| \dot{\va } \|_2 \| c_i \vb^i \|_2 ( \cos \psi_i - \cos \zeta \cos \varphi_i).
    \end{equation*}
    Note that $\lim_{t \rightarrow T^*} \cos \psi_i=1$ for $i \in C_1$. So for every $\varepsilon>0$, there exists $\delta >0$ such that  
    \begin{equation*}
        1-\varepsilon \leq \cos \psi_i \leq 1, \quad  t \in (T^*-\delta,T^*).
    \end{equation*}
    Set $\bar{t}_i = T^* -\delta$ and $\bar{\theta}_i = \arccos (1- \varepsilon)$. Then we have  either $\cos \varphi_i \ge \cos  \bar{\theta}_i, \ t \in (\bar{t}_i,T^*)$, or there exists $t \in (\bar{t}_i,T^*)$ such that $\cos \varphi_i \le \cos \bar{\theta}_i$, then it will increase monotonically until it goes up to $\bar{\theta}_i$. No matter in which case, we can find $\Tilde{\theta}_i \in [0, \frac{\pi}{2})$ such that $\cos \varphi_i \ge \cos \Tilde{\theta}_i$. Let $T_1 = \max_{i \in C_1} \bar{t}_i$ and $\Theta_1 = \max_{i \in C_1}  \Tilde{\theta}_i$. Thus, for each $i \in C_1$, the following inequality holds 
    \begin{equation*}
        \cos \varphi_i \ge \cos \Theta_{1},\quad t \in (T_1 ,T^*).
    \end{equation*}
    This completes the proof. 
\end{proof}

In fact, Proposition \ref{prop::4}  provides the angular relationship between 
$\va$ and its derivative $\dot{\va}$. We summarize it as follows.
\begin{prop}\label{prop::5}
Suppose that Assumption \ref{assump::FinalStageCond} holds. Consider the effective dynamics \eqref{eq::SimpEffectiveModel}.
    There exists $ T_2 \in (T_1,T^*)$ and $ \Theta_2 \in [0,\frac{\pi}{2})$ such that 
    \begin{equation}
        \cos \zeta \ge \cos \Theta_2, \quad t \in (T_2,T^*).
    \end{equation}
    Moreover, recall the definition of energy $E$, the following inequality holds
    \begin{equation}
        E= \langle \va, \dot{\va} \rangle \ge \cos \Theta_{2} \|\va \|_2 \|\dot{\va} \|_2 , \quad t \in (T_2,T^*).
    \end{equation}
\end{prop}

\begin{proof}
By definition of $\zeta$, we obtain
    \begin{equation*}
        \cos \zeta = \frac{\langle \dot{\va},\va  \rangle }{ \|\va \|_2 \| \dot{\va} \|_2 } =\frac{\langle \sum_{j=1}^m c_i \vb^i,\va  \rangle }{ \|\va \|_2 \| \dot{\va} \|_2 }.
    \end{equation*}
According to  Proposition \ref{prop::4}, we have 
\begin{equation*}
    \cos \zeta  \ge \cos \Theta_1 \frac{\sum_{i \in C_1} \|c_i \vb^i \|_2 }{\| \dot{\va} \|_2}, \quad t \in (T_1,T^*).
\end{equation*}
    Since we have shown that $\lim_{t \rightarrow T^*} \frac{\sum_{i \in C_1} \|c_i \vb^i \|_2}{\| \dot{\va} \|_2}=1 $ according to the proof of Proposition \ref{prop::Angle}, there exists $ T_2 \in (T_1,T^*)$ and $\Theta_2 \in [0, \frac{\pi}{2})$ such that 
    \begin{equation*}
        \cos \zeta \ge \cos \Theta_2 , \quad t \in (T_2,T^*).
    \end{equation*}
    Recall the definition of energy $E$, we have 
    \begin{equation*}
        E= \langle \va, \dot{\va} \rangle \ge \cos \Theta_{2} \|\va \|_2 \|\dot{\va} \|_2 , \quad t \in (T_2,T^*).
    \end{equation*}
    This completes the proof.
\end{proof}

Now we prove the theorem with all preparations above.

\begin{thm}[energy upper bound and blow up estimate]
\label{thm::UpperBound}
Suppose that Assumption \ref{assump::FinalStageCond} holds. Consider the effective dynamics \eqref{eq::SimpEffectiveModel}. There exist $T_3 \in (T_2,T^*)$ and $C \ge 1$ such that the following upper bound of Energy $E$ holds 
\begin{equation}
E(t) \le \frac{1}{\left( E(s)^{-\frac{1}{3}} - C(t-s) \right)^3}, \quad T_3 \le s < t < T^*.
\end{equation}
Moreover, the blow up time $T^*$ is bounded below by 
\begin{equation}
T^* \ge t + \frac{E(t)^{-\frac{1}{3}}}{C}, \quad t < T^*.
\end{equation}
\end{thm}

\begin{proof}
    First, by calculating the derivative of energy $E$, we have
    \begin{equation*}
        \dot{E} =\| \dot{\va} \|_2^2 + \| \dot{\vc} \|_2^2 + \| \va \|_2^2 \| \vc \|_2^2.
    \end{equation*}
    We rewrite the derivative of energy $E$ as 
    \begin{equation*}
        \begin{aligned}
            \dot{E} &= \| \dot{\va} \|_2^2 \left( 1+\frac{\|\dot{\vc} \|_2^2}{\| \dot{\va} \|_2^2} + \frac{\|\va\|_2^2 \|\vc \|_2^2}{\|\dot{\va}\|_2^2} \right) \\
        &=\| \dot{\va} \|_2^{\frac{4}{3}} \|\va \|_2^\frac{4}{3} \frac{\| \dot{\va} \|_2^\frac{2}{3}}{\| \va \|_2^\frac{4}{3}} \left( 1+\frac{\|\dot{\vc} \|_2^2}{\| \dot{\va} \|_2^2} + \frac{\|\va\|_2^2 \|\vc \|_2^2}{\|\dot{\va}\|_2^2} \right).
        \end{aligned}
    \end{equation*}
    According to  conservation laws and Equation (\ref{eq::AdotVsCnorm}), there exists $T_3>T_2$  such that 
    \begin{equation*}
        \frac{\| \dot{\va} \|_2^\frac{2}{3}}{\| \va \|_2^\frac{4}{3}} \left( 1+\frac{\|\dot{\vc} \|_2^2}{\| \dot{\va} \|_2^2} + \frac{\|\va\|_2^2 \|\vc \|_2^2}{\|\dot{\va}\|_2^2} \right) \le 4.
    \end{equation*}
    Then we have 
    \begin{equation*}
    \dot{E} \le 4 \| \dot{\va} \|_2^{\frac{4}{3}} \| \va \|_2^\frac{4}{3} ,\quad \forall t >T_3 .    
    \end{equation*}
    Based on Proposition \ref{prop::5}, we have 
    \begin{equation*}
        \| \va \|_2 \| \dot{\va} \|_2 \le \frac{1}{ \cos \Theta_{2}} E ,\quad \forall t> T_3.
    \end{equation*}
    Thus we obtain
    \begin{equation*}
        \dot{E} \le 4(\frac{1}{\cos \Theta_{2}})^\frac{4}{3}  E, \quad \forall t>T_3.
    \end{equation*}
    We denote $4 (\frac{1}{\cos \Theta_{2}})^\frac{4}{3} $ as $C_0$. Then we have $C_0 \ge 4$ and 
    \begin{equation*}
        \dot{E} \le C_0 E^\frac{4}{3}.
    \end{equation*}

    Opposite to proof of Theorem \ref{thm::BlowUp}, we obtain  
    \begin{equation*}
        \frac{\D}{\D t} E^{-\frac{1}{3}} \ge -\frac{1}{3} C_0,\quad  \forall t>T_3.
    \end{equation*}
    We denote $\frac{C_0}{3}$ as $C$. Thus $C \ge 1$ and we have 
    \begin{equation}
        E(t) \le \frac{1}{\left( {E(s)}^{-\frac{1}{3}}-C (t-s) \right)^3} , \quad  T_3 < s < t < T^*.
    \end{equation}
    Hence, for each time $t < T^*$, the time of blow up is bounded below by  
    \begin{equation}
        T^* \ge t + \frac{E(t)^{-\frac{1}{3}}}{C}.
    \end{equation}
    This completes the proof.
\end{proof}

\begin{thm}[condensation]
\label{thm::Condense}
Suppose that Assumption \ref{assump::FinalStageCond} holds. Dynamical system \eqref{eq::SimpEffectiveModel} has condensed solution.
\end{thm}

\begin{proof}
    We just prove the case for $c_i>0$. And the case for $c_i < 0 $ follows by similar argument. Because of  the derivative of $c_i$ is ${\vb^i}^{\T} \va$,  we only need to show that
    \begin{equation}
        \int_{T_3}^{T^*} {\vb^i}^{\T} \va \diff{t} = + \infty.
    \end{equation}
    Since $\lim_{t \rightarrow T^*} \frac{\|\va \|_2^2}{\|\dot{\va} \|_2}=1$ due to the statement (\romannumeral 3) of  Proposition \ref{prop::Angle}, there exists $T_4>T_3$ such that $\| \dot{\va} \|_2 \le 2 \sqrt{2} \| \va \|_2^2$, which implies
    \begin{equation}
    \label{eq::AdotA}
        E = \langle \dot{\va} , \va \rangle \le \norm{\dot{\va}}_2 \norm{\va}_2 \le 2 \sqrt{2} \norm{\va}_2^3.
    \end{equation}
    
    The idea is we can find a infinite division of $(T_4,T^*)$ such that the integral of ${\vb^i}^{\T} \va$ on each sub-interval is larger than positive constant which is an independent constant. 
 Then we consider the integral $\int_{t_1}^{t_2} {\vb^i}^{\T} \va \diff{t}$. By direct calculation of  the derivative of ${\vb^i}^{\T} \va$, we have 
    \begin{equation*}
        \dot{({\vb^i}^{\T} \va )}=c_i \va^{\T} \va + {\vb^i}^{\T} \left(\sum_j c_j \vb^j \right).
    \end{equation*}
    Integrating both sides of the equality, we have
    \begin{equation*}
        ({\vb^i}^{\T} \va) (t) = ({\vb^i}^{\T} \va) (t_1) + \int_{t_1}^{t} c_i \va^{\T} \va + {\vb^i}^{\T} \left(\sum_j c_j \vb^j \right) \D s
        \ge c_i(T_4) \int_{t_1}^{t} \va^{\T} \va \diff{s}.
    \end{equation*}
    Note that Equation \eqref{eq::AdotA} implies
    \begin{align*}
     \int_{t_1}^{t} \va^{\T} \va \diff{s} & \ge \frac{1}{ 2} \int_{t_1}^t E^{\frac{2}{3}} (s) \diff{s} \\
     &\ge \frac{1}{2} \int_{t_1}^t \frac{1}{(E(t_1)^{-\frac{1}{3}}-(s-t_1))^2} \diff{s} \\
    &=\frac{1}{2} \left[ \frac{1}{E(t_1)^{-{\frac{1}{3}}}-(t-t_1)} - \frac{1}{E(t_1)^{-\frac{1}{3}}} \right].
    \end{align*}
    Thus the integral satisfies 
    \begin{align*}
    \int_{t_1}^{t_2} {\vb^i}^{\T} \va \diff{t} & \ge c_i(T_4)  \frac{1}{2} \int_{t_1}^{t_2} \frac{1}{E(t_1)^{-{\frac{1}{3}}}-(t-t_1)} - \frac{1}{E(t_1)^{-\frac{1}{3}}} \diff{t} \\
    &= c_i(T_4) \frac{1}{2} \left[- \ln(E(t_1)^{-\frac{1}{3}} -(t_2 -t_1)) + \ln (E(t_1)^{-\frac{1}{3}}) -\frac{t_2-t_1}{E(t_1)^{-\frac{1}{3}}}\right].
    \end{align*}
    According to Theorem \ref{thm::UpperBound}, we can  choose $t_2 -t_1 =\frac{E(t_1)^{-\frac{1}{3}}}{ 2 C}$. Thus we obtain 
    \begin{equation}
        \int_{t_1}^{t_2} {\vb^i}^{\T} \va \diff{t} \ge c_i(T_4) \frac{1}{2} \left[\ln \frac{1}{1-\frac{1}{2 C}}-\frac{1}{ 2 C} \right] .
    \end{equation}
    Then we introduce auxiliary function 
    \begin{align*}
        f(t) &=\ln \frac{1}{1-t} -t \\
        &=- \ln (1-t) -t.
    \end{align*}
    We have $f(0)=0$ and $\dot{f}(t)>0$. Then we have $\ln \frac{1}{1-\frac{1}{2 C}}-\frac{1}{ 2 C} >0$. Since there are infinitely  many such sub-intervals, the proof of the theorem is completed.
\end{proof}



\appendix
\section{Final Stage Condition for Matrix Completion} 
\label{sec::FinalStageMC}

We will focus in this section on matrix sensing (a generalization of matrix completion) and discuss its connection and difference with initial condensation. Consider deep matrix factorization 
\begin{equation}
    \vW = \vW^{[3]} \vW^{[2]} \vW^{[1]},
\end{equation}
where $\vW^{[3]},\vW^{[2]},\vW^{[1]} \in {\sR}^{m \times m}$.  Its parameters are denoted  as  $\vtheta= \text{vec} \{ \vW^{[3]} , \mW^{[2]} ,\mW^{[1]} \}$ by flattening $\vW^{[3]}$, $\vW^{[2]}$ and $\vW^{[1]}$ and concatenating them together. Here, sample set $\vS$ is given by $n$ measurement matrices $\vX_1,\dots,\vX_n$ with corresponding labels $y_1,\dots,y_n$ generated by $y_i = \langle \vX_i, \vW^*  \rangle$ and out goal is to reconstruct $\vW^*$. The empirical risk to be minimized is 
\begin{equation}
    {\mR}_{\mS}(\vtheta)=\frac{1}{2n} \sum_{i=1}^n ( \langle \vW , \vX_i \rangle-y_i)^2 .
\end{equation}
We use gradient descent method (GD) to optimize the empirical risk. The parameters are initialized as follows:
\begin{equation}
    \vW_{kk'}^{[3]} \sim \fN (0,\varepsilon^2) ,  {\mW}_{kk'}^{[2]} \sim \fN (0,\varepsilon^2), {\mW}_{kk'}^{[1]} \sim \fN (0,\varepsilon^2), 
\end{equation}
where $\varepsilon$ is a small parameter.

Normalizing the parameters by 
\begin{equation*}
    \bar{\vW}^{[3]} = \varepsilon^{-1} \vW^{[3]} , \bar{\mW}^{[2]}= \varepsilon^{-1} \mW^{[2]}, \bar{\mW}^{[1]}= \varepsilon^{-1} \mW^{[1]}, \bar{\vtheta}= \varepsilon^{-1} \vtheta,
\end{equation*}
we have power series expansion of empirical risk in terms of  $\varepsilon$ as follows:
\begin{equation}
    \mR_{\mS} (\vtheta)= \frac{1}{2n} \sum_{i=1}^n y_i^2 - \varepsilon^3 \langle \bar{\vW}^{[3]} \bar{\mW}^{[2]} \bar{\mW}^{[1]} ,  \frac{1}{n} \sum_{i=1}^n y_i \vX_i \rangle  + o(\varepsilon^3).
\end{equation}
Then the leading order of GF obeys the following dynamics
\begin{equation}
    \varepsilon \frac{\D \bar{\vtheta}}{ \D t} =\varepsilon^2 \nabla_{\bar{\vtheta}} \langle \bar{\vW}^{[3]} \bar{\mW}^{[2]} \bar{\mW}^{[1]} ,  \frac{1}{n} \sum_{i=1}^n y_i \vX_i \rangle 
\end{equation}
Rescaling the time by setting
\begin{equation*}
    \bar{t}= \varepsilon  t
\end{equation*}
and dropping the bar in the above dynamics for simplicity, we obtain
\begin{equation*}
     \frac{\D \vtheta}{ \D t} = \nabla_{\vtheta} \langle \vW^{[3]} \mW^{[2]} \mW^{[1]} ,  \frac{1}{n} \sum_{i=1}^n y_i \vX_i \rangle.
\end{equation*}
Denote $\frac{1}{n} \sum_{i=1}^n y_i \vX_i $ as $\vV$ and write the dynamics into component form, 
\begin{equation}
    \left\{
    \begin{aligned}
    &\frac{\D \mW^{[3]}}{\D t} = \vV {\mW^{[1]}}^{\T} {\mW^{[2]}}^{\T} ,\\
    &\frac{\D \mW^{[2]}}{\D t}= {\vW^{[3]}}^{\T} \vV {\vW^{[1]}}^{\T},\\
    &\frac{\D \mW^{[1]}}{\D t}=  {\vW^{[2]}}^{\T} {\vW^{[3]}}^{\T} \vV.
    \end{aligned}
   \right.
\end{equation}
Assume that the measurement matrices $\vX_1,\dots,\vX_n$ are symmetric and commutable. $\vV$ can be orthogonally diagonalized into $\vLambda$. That is $\vV = \vO \vLambda \vO^{\T}$, where $\vO$ is an orthorgonal matrix. Let $\Tilde{\vW}^{[i]} = \vO^{\T} \vW^{[i]} \vO$, we obtain
\begin{equation}
    \left\{
    \begin{aligned}
    &\frac{\D \Tilde{\mW}^{[3]}}{\D t} = \vLambda {{}\tilde{\mW}^{[1]}}^{\T} {{}\Tilde{\mW}^{[2]}}^{\T} ,\\
    &\frac{\D \Tilde{\mW}^{[2]}}{\D t}= {{}\Tilde{\mW}^{[3]}}^{\T} \vLambda {{}\Tilde{\mW}^{[1]}}^{\T},\\
    &\frac{\D \Tilde{\mW}^{[1]}}{\D t}=  {{}\Tilde{\mW}^{[2]}}^{\T} {{}\Tilde{\mW}^{[3]}}^{\T} \vLambda.
    \end{aligned}
   \right.
\end{equation}
Let ${{}\Tilde{\mW}^{[3]}}^{\T} = (\va_1, \dots,\va_m)$, ${{}\Tilde{\mW}^{[1]}} = (\vc_1,\dots,\vc_m) $ and ${{}\Tilde{\vW}^{[2]}} = \vb$. We have 
\begin{equation}
\label{eq::MCEffDyn}
    \left\{
    \begin{aligned}
    &\frac{\D \va_i}{\D t} = \lambda_i \vb \vc_i ,\\
    &\frac{\D \vb }{\D t}=  \sum_{i=1}^m \lambda_i \va_i \vc_i^{\T} ,\\
    &\frac{\D \vc_i}{\D t}= \lambda_i \vb^{\T} \va,
    \end{aligned}
   \right.
\end{equation}
where $\lambda_i = \vLambda_{ii}$.

By introducing energy $E$ as
\begin{equation}
    E = \sum_{i=1}^m \lambda_i \va_i^{\T} \vb \vc_i,
\end{equation}
dynamical system $\eqref{eq::MCEffDyn}$ can be taken as gradient ascent of energy $E$. Without loss of generality, we assume $\lambda_i \neq 0$. For the case $\lambda_i = 0$ for some $i$, we neglect those terms.  Based on similar philosophy to our discussion about initial condensation, we generalize the concept of final stage condition.
\begin{assump}[final stage condition for matrix completion]
\label{assump::MCFSC}
    The parameters satisfy the final stage condition at time $t$. That is 
    \begin{enumerate}
        \item For every index $i \in [n]$ , $j \in [m]$ , we have $ \lambda_i c_{i,j} \va_i^{\T} \vb^j >0$ and $ \lambda_i a_{i,j} \vb_j \vc_i >0$.
        \item For every index $i \in [n] $ , $j,k \in [m]$, we have $ \langle c_{i,k} \vb^k, c_{i,j} \vb^j \rangle > 0 $ and $ \langle a_{i,k} \vb_k, a_{i,j} \vb_j \rangle > 0$.
        \item For every index $i,j \in [n]$ , $k,l \in [m]$, we have $ \lambda_i \lambda_j a_{i,k} a_{j,k} c_{i,l} c_{j,l} > 0$. 
    \end{enumerate}
\end{assump}

\begin{prop}[induction]
\label{prop::MCInduction}
Consider dynamical system \eqref{eq::MCEffDyn}. If Assumption \ref{assump::MCFSC} holds at some time $t_0 < T^*$, then Assumption \ref{assump::MCFSC}  will hold at $t_0 < t <T^*$. 
\end{prop}

\begin{proof}
First, the third condition means $c_{i,k} c_{j,k} c_{i,l} c_{j,l} > 0$ and $a_{i,k} a_{j,k} a_{i,l} a_{j,l} > 0$. This directly follows by direct calculation and symmetry:
\begin{equation}
    c_{i,k} c_{j,k} c_{i,l} c_{j,l} = \frac{1}{\lambda_i^2 \lambda_j^2 a_{i,k}^2 a_{j,k}^2 }(\lambda_i \lambda_j a_{i,k} a_{j,k} c_{i,k} c_{j,k})   (\lambda_i \lambda_j a_{i,k} a_{j,k} c_{i,l} c_{j,l})  >0 .
\end{equation}

    Calculating the derivative of the first condition, we have
\begin{equation}
    \frac{\D}{\D t} \lambda_i c_{i,j} \va_i^{\T} \vb^j= \dot{c}_{i,j}^2 +  \sum_{k=1}^m \lambda_i \lambda_k \va_i^{\T} \va_k c_{k,j} c_{i,j} + \lambda_i^2 \sum_{k=1}^m c_{i,k} {\vb^{k}}^{\T} \vb^{j} c_{i,j} >0
\end{equation}
and
\begin{equation}
    \frac{\D}{\D t} \lambda_i a_{i,j} \vb_j \vc_i =\dot{a}_{i,j}^2 + \sum_{k=1}^m \lambda_i \lambda_k a_{i,j} a_{k.j} \vc_k^{\T} \vc_i  + \lambda_i^2 \sum_{k=1}^m a_{i,j} \vb_j \vb_k^{\T} a_{i,k} > 0. 
\end{equation}
Then calculating the derivative of the second condition, we get 
\begin{equation}
\begin{aligned}
    \frac{\D}{\D t} \left( c_{i,k} {\vb^{k}}^{\T} \vb^{j} c_{i,j} \right) & = \left( c_{i,k} {\vb^{k}}^{\T} \vb^{j} c_{i,k} \right) \left(  \frac{1}{c_{i,k}^2} \left(\lambda_i c_{i,k} {\vb^{k}}^{\T} \va_i \right) +  \frac{1}{c_{i,j}^2} \left(\lambda_i c_{i,j} {\vb^{j}}^{\T} \va_i \right) \right)  \\
    & + \sum_{l=1}^m c_{i,k} c_{l,k} c_{i,j} c_{l,j} \left( \frac{1}{c_{l,j}^2}  \left(\lambda_l c_{l,j} {\vb^{j}}^{\T} \va_l \right) + \frac{1}{c_{l,k}^2}  \left(\lambda_l c_{l,k} {\vb^{k}}^{\T} \va_l \right) \right) > 0. 
\end{aligned}
\end{equation}
and another term follows by symmetry.
For the third condition, we have 
\begin{equation}
\begin{aligned}
    \frac{\D}{\D t}  \lambda_i \lambda_j a_{i,k} a_{j,k} c_{i,l} c_{j,l} &= \lambda_i \lambda_j a_{i,k} a_{j,k} c_{i,l} c_{j,l} \left( \frac{1}{a_{i,k}^2} \lambda_i a_{i,k} \vb_{k} \vc_i + \frac{1}{a_{j,k}^2} \lambda_j a_{j,k} \vb_{k} \vc_j \right)  \\ 
    &+  \lambda_i \lambda_j a_{i,k} a_{j,k} c_{i,l} c_{j,l} \left( \frac{1}{c_{i,l}^2} \lambda_i c_{i,l} {\vb^{l}}^{\T} \va_i + \frac{1}{c_{j,l}^2} \lambda_j c_{j,l} {\vb^{l}}^{\T} \va_j \right) >0. 
\end{aligned}
\end{equation}
This finishes the induction.
\end{proof}

\section*{Acknowledgments}
This work is sponsored by the National Key $\mathrm{R} \& \mathrm{D}$ Program of China Grant No. 2022YFA1008200
(T. L.), the National Natural Science Foundation of China Grant No. 12101401 (T. L.), Shanghai
Municipal Science and Technology Key Project No. 22JC1401500 (T. L.), Shanghai Municipal of
Science and Technology Major Project No. 2021SHZDZX0102, and the HPC of School of Mathematical Sciences and the Student Innovation Center, and the Siyuan-1 cluster supported by the Center
for High Performance Computing at Shanghai Jiao Tong University. The authors thank Zhi-Qin John Xu for helpful discussions. The authors also thank  OpenAI's ChatGPT for its aid in grammar checking and text refinement.

\bibliographystyle{siamplain}
\bibliography{references}
\end{document}